\pdfoutput=1

\documentclass[11pt]{article}

\usepackage[final]{acl}

\usepackage{times}
\usepackage{latexsym}

\usepackage[T1]{fontenc}

\usepackage[utf8]{inputenc}

\usepackage{microtype}

\usepackage{inconsolata}

\usepackage{graphicx}

\usepackage{algorithm}
\usepackage{algorithmic}
\usepackage{newfloat}
\usepackage{bbm}
\usepackage{amsmath}
\usepackage{multirow}
\usepackage{caption}
\usepackage{colortbl}
\usepackage{booktabs}
\usepackage{array}
\usepackage{listings}
\usepackage{tcolorbox}
\usepackage{multicol}
\usepackage{lipsum}
\usepackage{tabularx}
\usepackage{enumitem}
\usepackage{tocloft}
\usepackage{arydshln}
\PassOptionsToPackage{dvipsnames}{xcolor} 
\usepackage{xcolor} 

\title{Revitalizing Saturated Benchmarks: A Weighted Metric Approach for Differentiating Large Language Model Performance}

\author{
 \textbf{Bryan Etzine\textsuperscript{1*}},
 \textbf{Masoud Hashemi\textsuperscript{2}},
 \textbf{Nishanth Madhusudhan\textsuperscript{2}},
 \textbf{ Sagar Davasam\textsuperscript{2}},
\\
 \textbf{Roshnee Sharma\textsuperscript{2}},
 \textbf{Sathwik Tejaswi Madhusudhan\textsuperscript{2}},
 \textbf{Vikas Yadav\textsuperscript{2}},
\\
\\
 \textsuperscript{1}University of Florida,
 \textsuperscript{2}ServiceNow Research
\\
 \small{
  \texttt{bryetz@gmail.com, \{masoud.hashemi,nishanth.madhusudhan,sagar.davasam}}\\
\small{\texttt{
   roshnee.sharma,sathwiktejaswi.madhusudhan,vikas.yadav\}@servicenow.com
 }}
}

\begin{document}
\addtocontents{toc}{\protect\setcounter{tocdepth}{-1}}

\maketitle

\renewcommand{\thefootnote}{}
\footnotetext{\textsuperscript{*} Work done during internship in ServiceNow}
\renewcommand{\thefootnote}{\arabic{footnote}}

\begin{abstract}

Existing benchmarks are becoming saturated and struggle to separate model performances due to factors like data contamination and advancing LLM capabilities. This paper introduces \textit{EMDM} (Enhanced Model Differentiation Metric), a novel weighted metric that revitalizes benchmarks by enhancing model separation. EMDM integrates final answer and Chain-of-Thought (CoT) reasoning correctness, assigning weights based on the complexity and reasoning depth required to solve a given sample in the evaluation data. Using a \textit{baseline LLM} in two setups—``Unguided,'' where the model has no prior exposure to test samples, and ``Guided,'' where the model has prior knowledge of the desired answer—EMDM distinguishes instances of varying difficulty. The CoT and answer correctness from these setups inform an optimization objective for weight assignment, resulting in a more nuanced evaluation of model performance. Compared to the exact match (EM) metric, which achieves 17\% separation on ARC-Challenge \citep{clark2018think}, EMDM achieves 46\%, demonstrating its effectiveness in differentiating models based on reasoning and knowledge requirements.
\end{abstract}
\vspace{-3mm}
\begin{figure*}[t]
\centering
\includegraphics[width=0.85\textwidth]{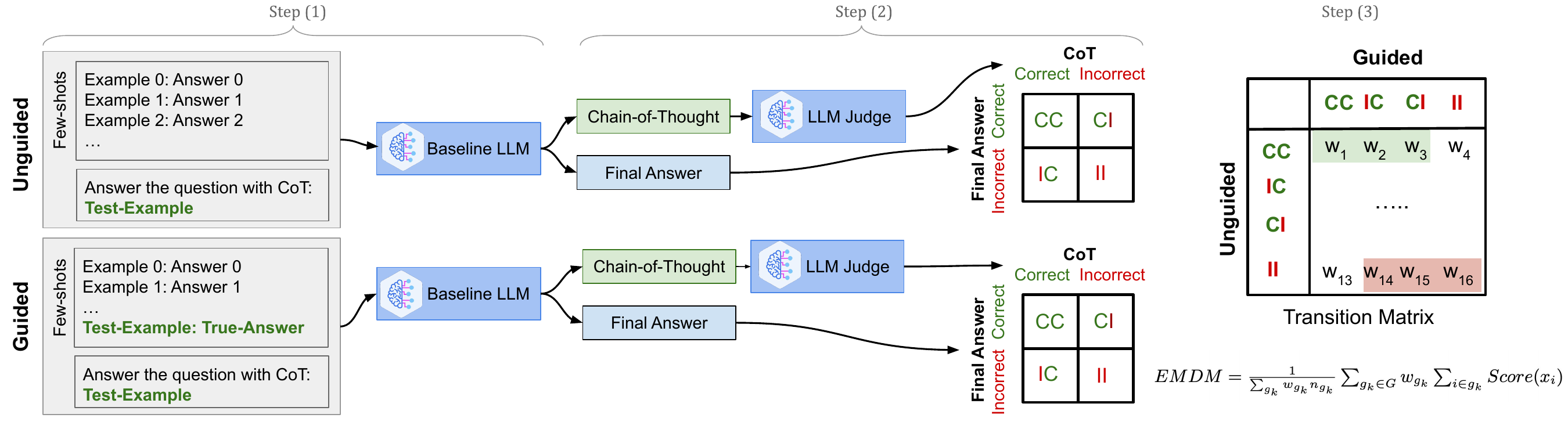}
\caption{\footnotesize{``Enhanced Model Differentiation Metric (EMDM)'' -- for each benchmark a baseline LLM is used to 1) generate guided and unguided responses, 2) create data categories based on the correctness of the answer and CoT facts and reasoning (the transition matrix), and 3) assign weights to each of the categories ($w_{g_k}$, see section \ref{sec:methods}) and calculate the weighted average.}}
\label{fig:metric_flow}
\vspace{-2mm}
\end{figure*}

\section{Introduction}

Large Language Models (LLMs) have made significant strides in NLP tasks \citep{qin2024large}, with benchmarks like MMLU \citep{hendrycks2020measuring}, ARC \citep{clark2018think}, and GSM8K \citep{cobbe2021training} playing a crucial role in tracking progress. However, model performance on these benchmarks has plateaued \citep{park2024open}, limiting their usefulness in differentiating models. Creating more complex benchmarks is resource-intensive and prone to inaccuracies in synthetic data generation \citep{zeng2024automatic, zeng2024skywork}.

We introduce the \textbf{Enhanced Model Differentiation Metric (EMDM)}, a novel approach to revitalizing existing benchmarks. While Chain-of-Thought (CoT) reasoning \citep{wei2022chain} has proven beneficial for fine-tuning \citep{mitra2023orca, selfthought2024}, its potential in evaluation is underutilized. EMDM addresses this by combining CoT correctness with traditional metrics like exact match (EM) and accuracy, emphasizing challenging examples that require deep reasoning.

Our methodology, illustrated in Figure \ref{fig:metric_flow}, uses a \textit{baseline LLM} with in-context learning (ICL) to generate responses under two setups: a \textit{Guided prompt} (with the test example and the desired response included, simulating contamination) and an \textit{Unguided prompt} (standard ICL). If a model fails under the Guided setup, it indicates limited reasoning for that sample. EMDM assigns higher weights to such challenging instances as they could be more differentiating (red highlights) and lower weights to easier ones (green highlights). Weights are calculated once per benchmark using a baseline LLM and seamlessly integrated into the evaluation process by updating only the final scoring.

\noindent
Our contributions in this paper include:
\begin{itemize}[noitemsep, topsep=0pt, left=2pt]
\item We introduce EMDM, a weighted metric combining answer correctness and CoT reasoning to emphasize challenging test instances, enhancing model differentiation.
\item We present an optimization objective to maximize benchmark score separation among LLMs. Our experiments demonstrate the effectiveness of the optimized weights in increasing the LLM performance differentiation. 
\end{itemize}

\section{Background \& Related Work} \vspace{-1mm}
In-Context Learning (ICL) allows LLMs to adapt to new tasks by showing a few relevant examples within the input prompt, without the need for further training or fine-tuning \citep{xu2024context}.
In practice, ICL is broadly applied for task learning (acquiring new skills) and task retrieval (activating pre-trained knowledge) \citep{brown2020language, lin2024dual}. We use ICL to prompt the model with two \textit{Guided} and \textit{Unguided} setups, showing task examples and providing direct guides in the prompts for the model to enhance its responses.

Chain of Thought (CoT), introduced by \citep{wei2022chain}, has improved LLM performance through step-by-step reasoning. While CoT judgment has been successfully used in LLM post-training methods such as Reinforced Fine-Tuning (ReFT) \citep{luong2024reft} to generate preference data pairs, its role in evaluation remains unexplored. EMDM fills this gap by incorporating CoT quality into LLM evaluation.

Human annotation is time-consuming and un-scalable. LLM-based evaluation has been widely adopted \citep{zheng2024judging, thakur2024judging}, showing high agreement with human judgment \citep{thakur2024judging, zheng2024judging}. Moreover, \citep{zeng2024mrben} introduce MR-Ben, a high quality benchmark designed to assess the capability of LLMs to serve as a judge for CoT reasoning samples. Their analysis evaluates several state-of-the-art LLMs across diverse domains, including mathematics, physics, coding, and logical reasoning, etc., and reveal that GPT-4o and GPT-4 series models consistently outperform other LLMs as reliable evaluators of CoT correctness. Building upon these motivations, we use LLM based judges to evaluate the accuracy of CoT responses, generated by baseline models.

Some benchmarks, such as WildBench \citep{lin2024wildbench} and FLASK \citep{ye2023flask}, group dataset samples by difficulty using LLMs. However, the reliability of LLM-based difficulty classification remains largely unexplored. In contrast, EMDM categorizes samples based on baseline LLM performance accounting for both CoT reasoning and answer correctness, without directly classifying difficulty (see Appendix \ref{apx:flask_vs_emdm} for comparison of FLASK and EMDM).

\vspace{-2mm}
\section{EMDM Formulation} \vspace{-1mm}
\label{sec:methods}
EMDM enhances LLM performance separation through: (1) sample categorization and (2) weight optimization. The dataset samples are first categorized based on the correctness of the answer and the CoT judgment into four quadrants: (Correct/Incorrect Answer) $\times$ (Correct/Incorrect CoT). This procedure is applied to two prompting strategies with the baseline model (Figure \ref{fig:metric_flow}-step (1)):
\\ \noindent
\textbf{Unguided prompt}: Follows the standard ICL setup, where test samples or related information are excluded from few-shot prompts.
\\ \noindent
\textbf{Guided prompt}: Includes the test example and the desired answer in the few-shot prompts, simulating contamination and identifying cases where the model fails even with direct guidance and access to the correct answer.  

All responses generated by the two prompt types require CoT. The generated CoT and answers are then assessed by (Figure \ref{fig:metric_flow}-step (2)):
\\ \noindent
\textbf{Exact Match}: Checks whether the model's response matches the correct response, thereby assessing the accuracy of the final answer.
\\ \noindent
\textbf{LLM-Judge} \citep{zheng2024judging}: Assesses whether the reasoning steps effectively support the conclusion, regardless of the final answer's correctness (see Appendix \ref{apdx:prompts}). We use GPT-4 to critique the CoT generated by the baseline.

Using the above process, we generate \( 4 \times 4 = 16 \) categories (Figure \ref{fig:metric_flow}-step (3)), denoted by $g_k \in G$, in the form of a \textit{sample transition matrix}, rows being (Correct/Incorrect Answer) $\times$ (Correct/Incorrect CoT) from unguided prompt and columns being (Correct/Incorrect Answer) $\times$ (Correct/Incorrect CoT) from guided prompt. Each transition matrix cell (a.k.a., category) captures how samples transition between different quadrants based on the prompt type. This matrix offers insights into how the presence of strong guidance influences the model's ability to produce correct answers and reason effectively.
\\ \noindent
\textbf{Weight Optimization:} Weights, denoted by \( \hat{w} \), are calculated for each of the 16 categories to maximize the pairwise difference between the LLMs, considering all LLM pairs, \( A \) and \( B \), excluding the baseline model \( R \), by solving the following optimization:
\vspace{-2mm}
\begin{equation}
\small
\begin{split}
&\hat{w} = \mathop{\mathrm{argmax}}_{\substack{w=\{w_{g_k}\}_{k=1}^N \\ L \leq w_{g_k} \leq U}} 
-\sum_{g_k \in G} w_{g_k}  
+ \sum_{g_k \in G} \frac{w_{g_k}}{n_{g_k} \cdot \binom{|M|}{2}} \\
&\quad \times \sum_{\substack{(A, B) \in M \\ A \neq B \\ A, B \neq R}} \sum_{i \in g_k} \left| Score^A(x_i) - Score^B(x_i) \right|
\end{split}
\label{eqn:pairwise_weight_optimization}
\end{equation}
\vspace{-1mm}
\noindent where \( L \) and \( U \) are the lower and upper bounds of the weights, \( N \) is the number of categories (e.g., 16 in EMDM), model pairs (\(A, B\)) are chosen from a set of \( M \) LLMs, which can be the same as or different from the models being evaluated, \( Score \) is the evaluation score for the given input sample \( x_i \), e.g., exact match, and \( n_{g_k} \) is the number of samples in category \( g_k \in G \).

The first term regularizes the weights toward the lower bound $L$, while the second term, capturing pairwise score differences, pushes them toward the upper bound $U$. Therefore, the combined objective assigns higher weights to categories with larger score differences, due to their stronger gradients, and lower weights to those with smaller differences. This ensures that the weights highlight categories where the models exhibit the most variation, enhancing the differentiation between them.

These weights are calculated once and will be used in LLM benchmarking. Using the calculated weights for samples in each category (\( \hat{w}_{g_k} \)), EMDM is defined as a weighted average of the benchmark sample scores:
\vspace{-1mm}
\begin{equation*}
\footnotesize
\scalebox{1}{$
    EMDM = 
    \frac{1}{\sum_{g_k} \hat{w}_{g_k} n_{g_k}} \sum_{g_k \in G} \hat{w}_{g_k} \sum_{i \in g_k} Score(x_i)
    $}
\end{equation*}

\section{Experiments}
\textbf{Experiment set-up:}
We evaluate GPT-3.5 \citep{ouyang2022training}, GPT-4 \citep{achiam2023gpt}, GPT-4-Turbo, GPT-4o, Mixtral-8x7B-Instruct \citep{jiang2024mixtral}, Mixtral-8x22B-Instruct, Mistral 7B-Instruct, and Llama3.1-8b-Instruct \citep{dubey2024llama}\footnote{we discard ``Instruct'' from model names for brevity.} on MMLU \citep{hendrycks2020measuring}, ARC-Challenge \citep{clark2018think}, TruthfulQA \citep{lin2021truthfulqa}, and GSM8K \citep{cobbe2021training}. The first three are multiple choice question-answering (MCQA), while GSM8K involves free-form text generation. GPT-4 judges CoT reasoning correctness. For MCQA, the baseline model provides reasoning for each option, eliminating incorrect choices step-by-step. This process provides more information to judge the model. In GSM8K, we compare the model's CoT with the ground truth CoT given in the dataset. We follow the standard evaluation setup for each benchmark: e.g., 10-shot for ARC-Challenge and 5-shot for MMLU. We use $L=0.1$ and $U=2.0$ in the results presented here (more details in Appendix \ref{apdx:landu}), and optimization is solved using Scipy SLSQP (see \ref{apdx:linear_programming}) \citep{2020SciPy-NMeth}. Mistral 7B is used as the baseline LLM in the experiments (effect of baseline models on EMDM is studied in Appendix \ref{apx:baselines}).

\noindent\textbf{Ablation Study:} Three categorization results are reported: 1) EMDM (ours) with $4 (\texttt{Unguided}) \times 4 (\texttt{Guided}) = 16$ categories created based on correctness of the answer and CoT, 2) Answer-only with $2 (\texttt{Unguided}) \times 2 (\texttt{Guided}) = 4$ categories only based on the correctness of the answer, and 3) Unguided-only with 4 categories from answer and CoT correctness of unguided prompts.

\begin{figure}[h!]
    \centering
    \includegraphics[width=0.8\linewidth]{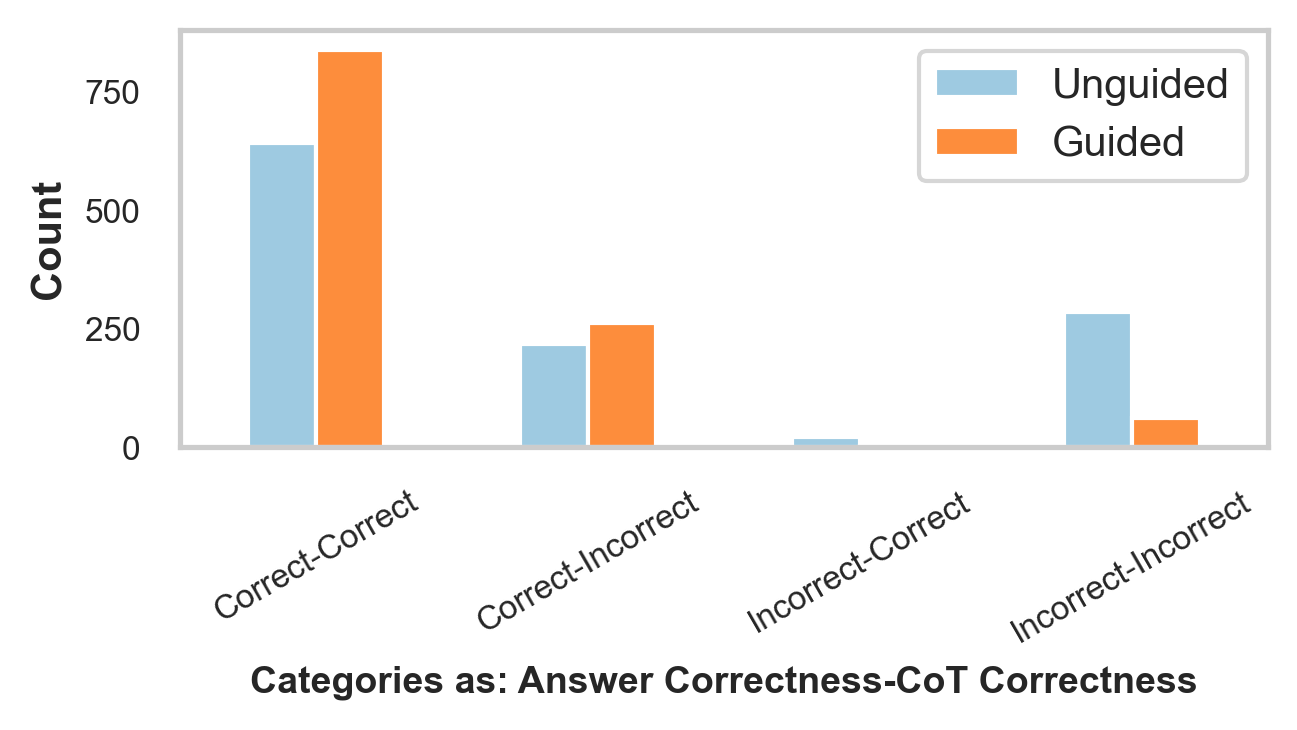}
    \caption{\footnotesize{ARC-Challenge sample distribution with Mistal 7B on Unguided \& Guided prompt responses.}}
    \label{fig:mistral_unguided_arc_c}
\end{figure}

\noindent\textbf{Guided vs. Unguided Sample Transition:}
Figure \ref{fig:mistral_unguided_arc_c} shows the distribution of samples across the four quadrants of the Answer/CoT correctness matrix in the ARC-Challenge benchmark. Including guided examples notably increases samples with Correct Answers and Correct CoT, as well as those with Correct Answers but Incorrect CoT. This suggests that guiding examples can lead to correct responses even if the underlying reasoning is flawed.
Table \ref{tab:transition_matrix} displays the transition of samples among quadrants after adding guidance to the ICL prompt. The most notable transitions include: 1) from Correct Answer \& Incorrect CoT to Correct Answer \& Correct CoT (137 samples), indicating improved reasoning with guidance; 2) from Incorrect Answer \& Incorrect CoT to Correct Answer \& Correct CoT (125 samples), showing enhanced accuracy and reasoning; and 3) from Incorrect Answer \& Incorrect CoT to Correct Answer \& Incorrect CoT (113 samples), where the model guesses the correct answer without accurate reasoning.

Table \ref{tab:transition_matrix} also shows the weights assigned to each category using Eq. \ref{eqn:pairwise_weight_optimization}. Samples correctly answered and reasoned in the unguided prompts (CC) have lower weights, while those incorrect in both answer and reasoning (II), even with guidance, are considered the most challenging and receive the highest weight of 2.0 (results with intuitive vs. optimized weight assignment are available in \ref{apdx:intuitive_weights}).

\begin{table}[ht]
\centering
\resizebox{0.7\columnwidth}{!}{
\begin{tabular}{@{}cc|cccc@{}}
 & & \multicolumn{4}{c}{\textbf{Guided}} \\
 & & \textbf{CC} & \textbf{CI} & \textbf{IC} & \textbf{II} \\ 
\midrule
\multirow{4}{*}{\rotatebox{90}{\textbf{Unguided}}} 
& \textbf{CC}  & \cellcolor{green!25}559 (0.1) & 67 (0.1) & 2 (1.0*) & 12 (0.8)  \\
& \textbf{CI}  & \cellcolor{gray!50} 137 (0.1) & 77 (0.1) & 0 (-) & 4 (1.0*)  \\
& \textbf{IC}  & 16 (0.1) & 6 (1.0*) & 0 (-) & 1 (1.0*)  \\
& \textbf{II}  & \cellcolor{gray!50}125 (0.1) & \cellcolor{gray!50}113 (2.0) & 1 (1.0*) & \cellcolor{red!25}46 (2.0)  \\
\bottomrule
\end{tabular}
}
\caption{\footnotesize{Transition matrix for ARC-Challenge. The first (C)orrect/(I)ncorrect indicates answer correctness, the second C/I represents CoT correctness. Values in parentheses are weights assigned to each category by Eq.\ref{eqn:pairwise_weight_optimization}. Cells with fewer than 10 samples are assigned a weight of 1.0, marked by an asterisk. Gray highlights show the cells with the highest counts. Green and Red highlights are to match figure \ref{fig:metric_flow}.}}
\label{tab:transition_matrix}
\end{table}

\noindent\textbf{EMDM Revitalizing Benchmarks:}
Table \ref{tab:emdm-benchmarks} compares EMDM with exact match (EM) across MMLU, ARC-Challenge, TruthfulQA, and GSM8K. The score range (gap between the baseline and the most powerful model) and the inter-model difference (model separation) increase across all benchmarks with EMDM. The largest increases are observed in ARC-Challenge and TruthfulQA where the score range jumps from 17.0\% and 41.2\% in EM to 46.8\% and 61.6\% in EMDM, and the average difference rises from 2.5\% and 5.9\% to 6.7\% and 8.81\% accordingly.
\\ \noindent
In GSM8K, while GPT-4 and GPT-4 Turbo achieve the same score under EM, EMDM reveals subtle differences, with GPT-4 Turbo outperforming on questions requiring more reasoning. In TruthfulQA, Llama3.1-8B outperforms Mixtral-8x7B by 4.80\% in EMDM, whereas Mixtral-8x7B was 1.05\% better in EM, suggesting that Mixtral-8x7B excels at answering simpler questions compared to Llama3.1-8B. Additionally, the separation between GPT-4o and both GPT-4-Turbo and GPT-4-32K increases significantly, though the gap between GPT-4-32K and GPT-4-Turbo narrows.\\ \noindent
In MMLU, Humanities (see Appendix \ref{apx:mmlu-results}) is the only subject showing improvement, while overall scores (All subjects) remain largely unchanged. Using a better baseline LLM can increase the separation in such cases (Appendix \ref{apx:baselines}).
\\ \noindent
\textbf{Ablation Results:} Comparing `Answer-only' with EMDM, in some cases, it offers a better separation than EMDM. Therefore, `Answer-only' could serve as a cheaper alternative to EMDM, as it does not require CoT judgments. However, on average, EMDM is more consistent across all benchmarks.

\begin{table}[h!]
\centering
\resizebox{\columnwidth}{!}{
\begin{tabular}{l|rrrr}
\toprule
{} & \textbf{EM} & \textbf{EMDM (ours)} & \textbf{Answer-only} & \textbf{Unguided-only} \\
\hline
\multicolumn{5}{c}{\textbf{ARC-Challenge}} \\
\hline
Mistral 7B         &   0.79 (0.00\%) &   0.46 (0.00\%) &   0.52 (0.00\%) &   0.49 (0.00\%) \\
Llama 3.1 8B       &   0.84 (6.01\%) &  0.55 (17.54\%) &  0.61 (13.65\%) &  0.60 (19.17\%) \\
GPT3.5             &   0.85 (7.88\%) &  0.60 (24.56\%) &  0.68 (22.96\%) &  0.65 (24.94\%) \\
Mixtral 8x7B       &   0.87 (8.97\%) &  0.63 (27.15\%) &  0.71 (25.58\%) &  0.68 (28.32\%) \\
Mixtral 8x22B      &  0.92 (14.30\%) &  0.76 (39.66\%) &  0.83 (36.51\%) &  0.80 (39.22\%) \\
GPT4-32K           &  0.95 (17.44\%) &  0.83 (44.99\%) &  0.92 (42.90\%) &  0.88 (44.72\%) \\
GPT4-Turbo         &  0.96 (17.59\%) &  0.85 (46.26\%) &  0.93 (43.52\%) &  0.89 (45.62\%) \\
GPT4o              &  0.95 (17.00\%) &  0.86 (\textbf{46.81}\%) &  0.92 (42.72\%) &  0.89 (45.37\%) \\
\hdashline
\textbf{Inter-Model Difference} &         +2.51\% &         \textbf{+6.69}\% &         +6.22\% &         +6.52\% \\
\hline
\multicolumn{5}{c}{\textbf{TruthfulQA}} \\
\hline
Mistral 7B         &   0.52 (0.00\%) &   0.30 (0.00\%) &   0.30 (0.00\%) &   0.36 (0.00\%) \\
GPT3.5             &  0.66 (21.75\%) &  0.47 (35.94\%) &  0.47 (36.40\%) &  0.54 (33.68\%) \\
Mixtral 8x7B       &  0.70 (26.14\%) &  0.48 (37.71\%) &  0.48 (37.91\%) &  0.58 (37.35\%) \\
Llama 3.1 8B       &  0.69 (25.09\%) &  0.52 (42.51\%) &  0.52 (42.29\%) &  0.59 (39.25\%) \\
Mixtral 8x22B      &  0.78 (33.70\%) &  0.66 (54.08\%) &  0.65 (54.27\%) &  0.71 (49.36\%) \\
GPT4o              &  0.82 (36.50\%) &  0.68 (55.64\%) &  0.68 (56.12\%) &  0.75 (51.78\%) \\
GPT4-Turbo         &  0.85 (39.16\%) &  0.77 (60.95\%) &  0.77 (61.33\%) &  0.80 (54.70\%) \\
GPT4-32K           &  0.88 (41.20\%) &  0.79 (61.65\%) &  0.78 (\textbf{62.01}\%) &  0.83 (56.35\%) \\
\hdashline
\textbf{Inter-Model Difference} &         +5.89\% &         +8.81\% &         \textbf{+8.86}\% &         +8.05\% \\
\hline
\multicolumn{5}{c}{\textbf{GSM8K}} \\
\hline
Mistral 7B         &   0.49 (0.00\%) &   0.31 (0.00\%) &   0.52 (0.00\%) &   0.33 (0.00\%) \\
GPT3.5             &  0.58 (15.03\%) &  0.46 (32.69\%) &   0.55 (5.24\%) &  0.49 (32.78\%) \\
Mixtral 8x7B       &  0.69 (28.34\%) &  0.54 (42.88\%) &  0.64 (18.49\%) &  0.57 (42.59\%) \\
Llama 3.1 8B       &  0.77 (36.02\%) &  0.64 (51.56\%) &  0.79 (34.64\%) &  0.67 (51.22\%) \\
Mixtral 8x22B      &  0.87 (43.43\%) &  0.79 (61.09\%) &  0.87 (40.30\%) &  0.82 (59.62\%) \\
GPT4-32K           &  0.93 (47.20\%) &  0.88 (65.07\%) &  0.95 (45.22\%) &  0.90 (63.37\%) \\
GPT4o              &  0.94 (47.58\%) &  0.90 (65.57\%) &  0.96 (46.01\%) &  0.91 (63.81\%) \\
GPT4-Turbo         &  0.94 (47.58\%) &  0.90 (\textbf{65.64}\%) &  0.93 (44.38\%) &  0.91 (63.78\%) \\
\hdashline
\textbf{Inter-Model Difference} &         +6.85\% &         \textbf{+9.38}\% &         +6.57\% &         +9.12\% \\
\hline
\multicolumn{5}{c}{\textbf{MMLU: All Subjects}} \\
\hline
Mistral 7B         &   0.67 (0.00\%) &   0.66 (0.00\%) &   0.68 (0.00\%) &   0.68 (0.00\%) \\
Llama 3.1 8B       &   0.74 (8.30\%) &   0.71 (7.12\%) &   0.74 (8.09\%) &   0.74 (7.04\%) \\
GPT3.5             &  0.75 (10.36\%) &   0.73 (9.89\%) &  0.76 (10.82\%) &   0.74 (7.84\%) \\
Mixtral 8x7B       &  0.77 (12.06\%) &  0.76 (13.45\%) &  0.78 (12.66\%) &  0.77 (11.21\%) \\
Mixtral 8x22B      &  0.82 (17.62\%) &  0.82 (19.08\%) &  0.83 (18.01\%) &  0.83 (17.61\%) \\
GPT4-32K           &  0.85 (21.04\%) &  0.86 (23.20\%) &  0.87 (21.53\%) &  0.87 (20.98\%) \\
GPT4-Turbo         &  0.86 (21.32\%) &  0.86 (23.34\%) &  0.87 (21.56\%) &  0.87 (21.38\%) \\
GPT4o              &  0.88 (23.28\%) &  0.88 (\textbf{25.20}\%) &  0.89 (23.73\%) &  0.89 (23.39\%) \\
\hdashline
\textbf{Inter-Model Difference} &         +3.33\% &         \textbf{+3.60}\% &         +3.39\% &         +3.34\% \\
\bottomrule
\end{tabular}
}
\caption{\footnotesize{Benchmarking Results -- The values in parentheses indicate the percentage increase in score relative to the baseline (Mistral 7B, represented as 0\%). `Inter-Model Difference' shows the average model separation (difference from the closest model).}}
\label{tab:emdm-benchmarks}
\end{table}

\vspace{-3mm}
\section{Conclusion}

We introduced the \textit{Enhanced Model Differentiation Metric (EMDM)}, a novel metric that enhances model differentiation through guided and unguided promptings. EMDM combines CoT judgment and answers correctness to assess reasoning depth. In the guided ICL setup, where the correct answer is included in the prompt, models do not always produce correct responses or reasoning paths (i.e., CoT), leading to varied behaviours across samples. EMDM leverages these differences to identify challenging instances and assign appropriate weights. Tested across benchmarks, EMDM improves performance separation, especially in saturated benchmarks. For difficult benchmarks, adjusting weight bounds and baseline model selection can further enhance separation (Appendix \ref{apdx:hyperparams}).

\section{Limitations}
\begin{itemize}[noitemsep, topsep=0pt, left=2pt]
\item A rigorous CoT judgment accuracy analysis is needed. LLM-based CoT evaluation may introduce bias, as the LLM-judge could have limitations or errors, potentially affecting weight assignment accuracy. However, studies are showing that LLMs are powerful judges \citep{zheng2024judging, thakur2024judging}. We tested our judge prompt using synthetically generated wrong CoTs, however, those CoTs do not meet the complexities observed in the real CoTs generated in benchmarks.
\item Potential biases in LLM-as-a-judge: Although we do not use a judge for a similar model, which reduces the potential of the self-favour, other biases such as lengthy response bias have not been studied in this paper.
\item We lacked access to proprietary models like Gemini and Claude. Exploring the effects of EMDM on these models and comparing their performance with GPT-4 could provide additional insights.
\item Lastly, we have not tested EMDM on benchmarks where exact match does not apply, leaving this for future work.
\end{itemize}

\section{Ethical Consideration}

In our experiments, we utilize popular evaluation datasets like ARC-Challenge, TruthfulQA, GSM8K, and, MMLU. There could be a few potentially sensitive questions in TruthfulQA as per the authors of \cite{lin2021truthfulqa}. We also utilize off-the-shelve LLMs without any finetuning from our end as our study is focused on revitalizing evaluation benchmarks with EMDM. We kindly refer readers to disclaimers of respective LLMs used in our experiments.

\bibliography{references}

\appendix
\section{Appendix}
\addtocontents{toc}{\protect\setcounter{tocdepth}{2}}
\renewcommand{\contentsname}{\footnotesize{Appendix Table of Contents}}
\tableofcontents

\begin{figure*}[th!]
\centering
\includegraphics[width=0.9\textwidth]{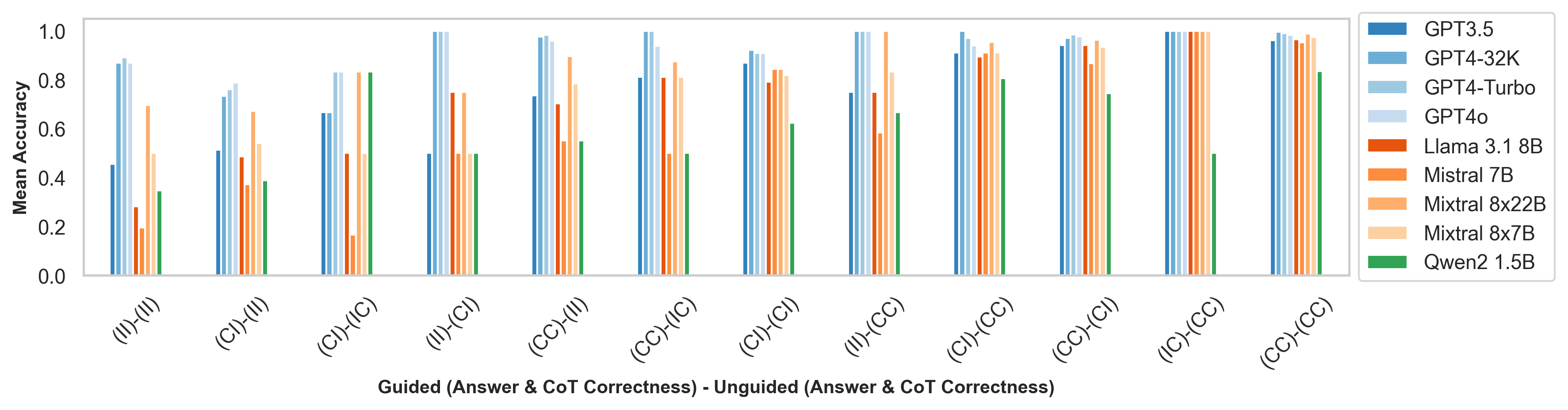}
\caption{\footnotesize{The average exact match (EM) accuracy in different sample groups of ARC-Challenge, with Mistral7B as the baseline. The groups with 0 or 1 sample are not shown.}}
\label{fig:arc_c_groups_acc}
\end{figure*}

\subsection{EMDM's Hyper-parameters}
\label{apdx:hyperparams}
EMDM has two hyper-parameters: 1) the baseline model and 2) the upper and lower bounds. In the following sections, we present some results showing the effect of the hyper-parameters on the EMDM results.

\subsubsection{Effect of the Baseline Model}
\label{apx:baselines}
One of the most important decisions to make in EMDM is the choice of the baseline. The role of the baseline is to identify the samples that all the models correctly answer with a high probability. Here we show the results of ARC-Challenge with different baselines: Qwen2-1.5B \citep{yang2024qwen2}, Mistral7B, and GPT3.5.
Table \ref{tab:arc_c_baseline_study} presents a comparison of scores from exact match (EM), EMDM, and scores obtained by ablating CoT correctness and guided prompting, using different baselines. When Qwen2-1.5B is the baseline, the score range in EMDM increases from 25.94\% in EM to 56.53\%. However, the separation among models, particularly within the GPT4 class, does not improve. Although the inter-model difference increases from 3.26\% to 7.07\%, this is primarily due to the significant increase in the separation of Mistral 7B (from 10.29\% to 36.32\%). This is a result of the performance gap between Qwen2-1.5B and other models.

Using Mistral 7B as the baseline, EMDM assigns a higher score to Qwen2-1.5B than Mistral7B. Figure \ref{fig:arc_c_groups_acc} helps to clarify this further. Examining the performance of different models across various Answer/CoT correctness categories reveals that the groups with low weights (calculated by Eq. \ref{eqn:pairwise_weight_optimization} -- see table \ref{tab:transition_matrix}) are those where all models, except Qwen2-1.5B, perform nearly perfectly. By down-weighting these samples, where Qwen2-1.5B underperforms compared to others, EMDM gives Qwen2-1.5B a score higher than it merits. A similar effect is seen with Llama3.1-8B when GPT3.5 is used as the baseline. When GPT3.5 is the baseline, the highest separation among the more powerful models, particularly the GPT4 class, is achieved. Using EM, there is no separation among the GPT4 family (0.16\% and 0.48\% -- insignificant differences), but when using EMDM with GPT3.5 as the baseline, a GPT4-Turbo and GPT4-32K are differentiated from GPT4o with a (3.21\% and 3.21+0.31\%).

\begin{figure}
    \centering
    \includegraphics[width=0.49\linewidth]{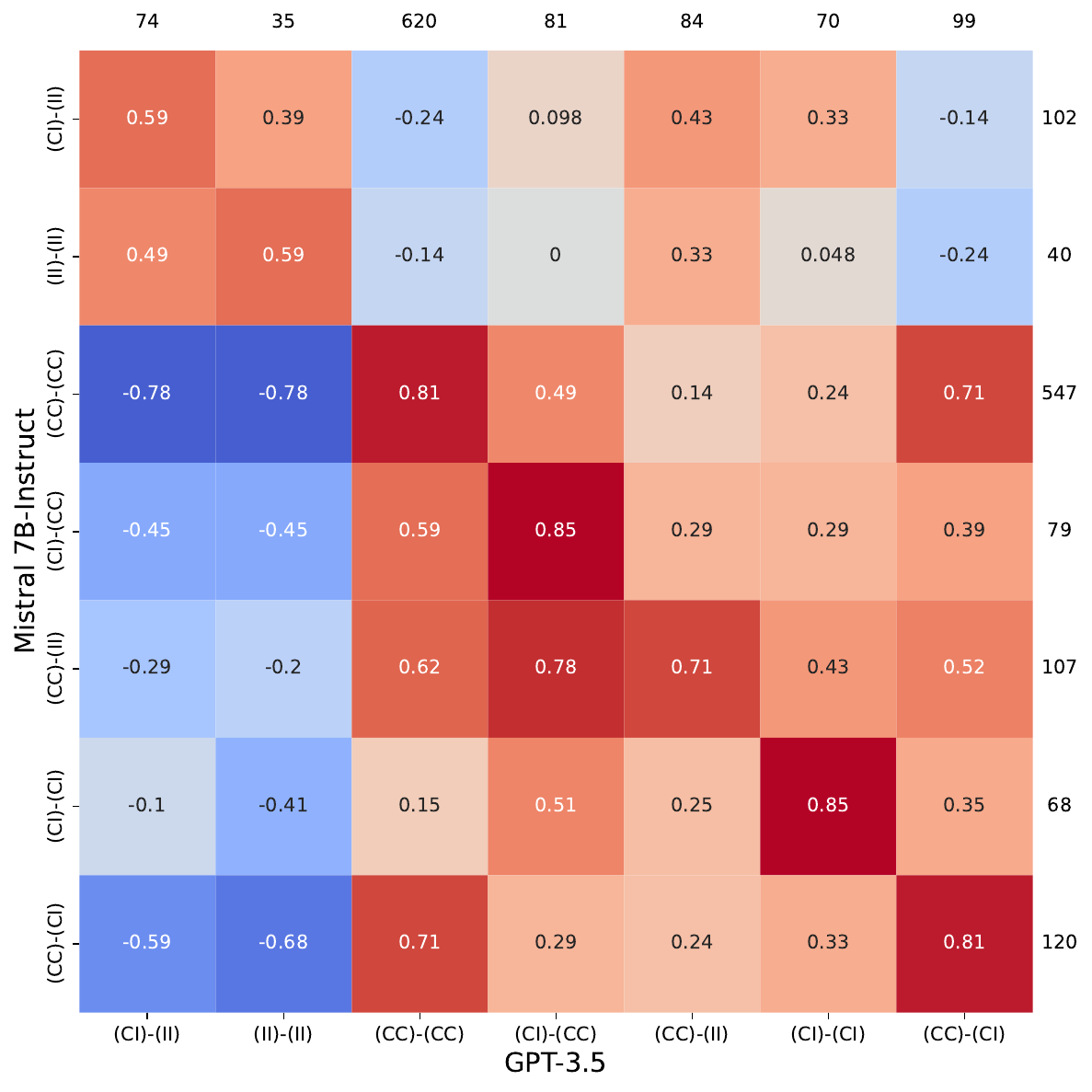}
    \includegraphics[width=0.49\linewidth]{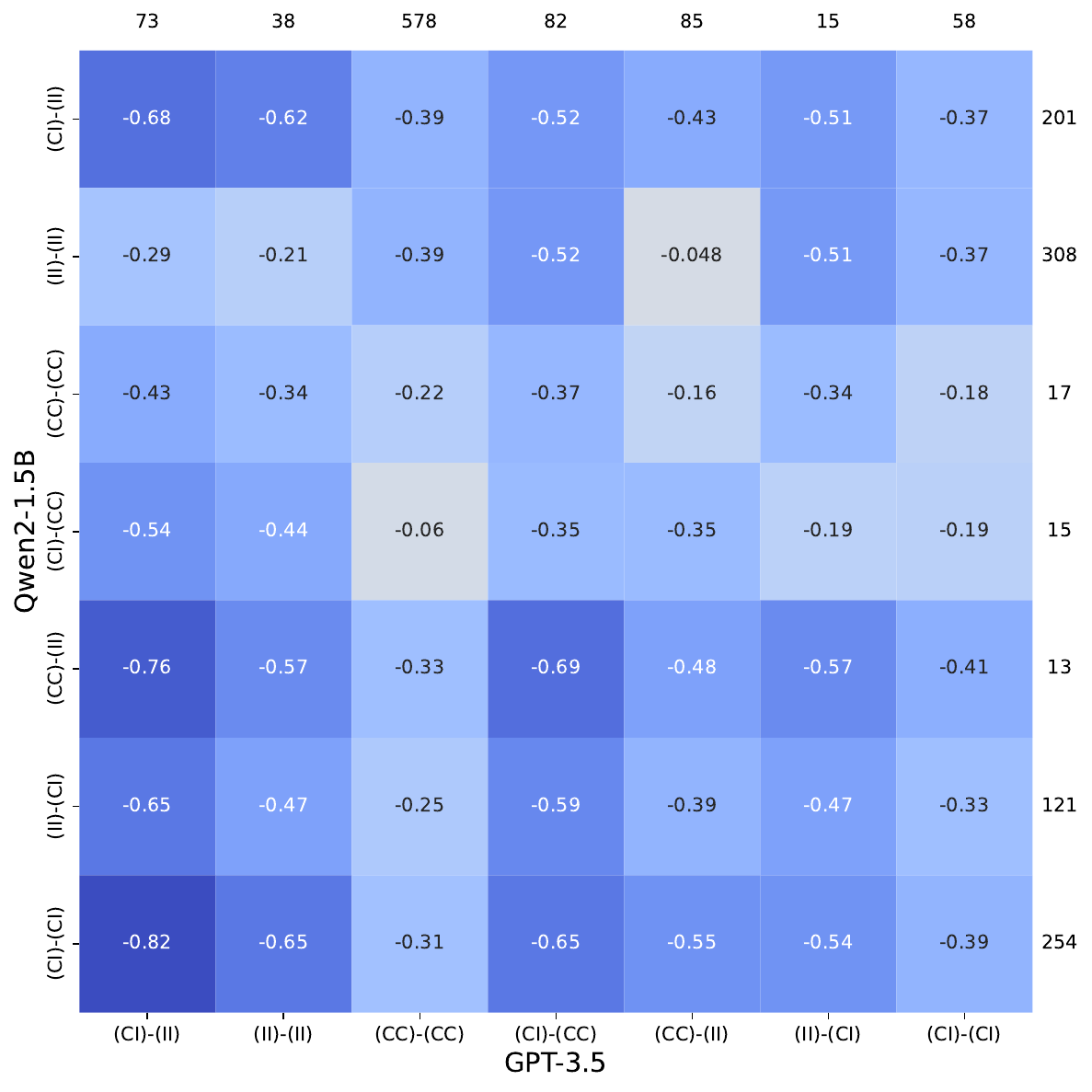}
    \caption{\footnotesize{Kendall's Tau correlation between (Left) GPT-3.5 and Mistral 7B-Instruct and (Right) Qwen2-1.5B and GPT3.5. The numbers on top and right show the marginal count of the samples in each category. Ones with less than 10 samples are removed (which means the margins may differ due to the removal of those cells). The cells are Guided (Answer Current/Incorrect, CoT Correct/Incorrect)-Unguided (Answer Correct/Incorrect, CoT Correct/Incorrect).}}
    \label{fig:baseline-heatmap}
\end{figure}

To understand the behaviour of the baselines comparatively, Figure \ref{fig:baseline-heatmap} shows the Kendall's Tau correlation between the categories of the baselines. Mistral 7B and GPT-3.5 show a very high correlation, while Qwen2-1.5B has almost no correlation with GPT-3.5. As can be seen from the figure, Qwen2-1.5 responses are mostly placed in (CI)-(CI), (II)-(II), (II)-(CI) and (CI-II), showing that its CoT quality is almost always low. This shows that smaller models cannot generate high quality CoTs even when their final responses are correct.

To achieve significant separation among models, the baseline should not be a low-performing model. Simultaneously, the baseline should be the lowest-scoring model among those being benchmarked. For the experiments in this paper, Mistral 7B is used as our baseline and we discard Qwen2-1.5B from those experiments. In an application such as fine-tuning a model iteratively, an early checkpoint can be used as the baseline.

\begin{table}[t!]
\centering
\resizebox{\columnwidth}{!}{
\begin{tabular}{l|rrrr}
\toprule
{} & \textbf{EM} & \textbf{EMDM (ours)} & \textbf{Answer-only} & \textbf{Unguided-only} \\
\hline
\multicolumn{5}{c}{\textbf{Qwen2 1.5B as Baseline}} \\
\hline
\rowcolor{gray!20} Qwen2 1.5B         &   0.71 (0.00\%) &   0.40 (0.00\%) &   0.39 (0.00\%) &   0.45 (0.00\%) \\
Mistral 7B         &  0.79 (10.29\%) &  0.64 (36.32\%) &  0.63 (37.15\%) &  0.66 (30.96\%) \\
Llama 3.1 8B       &  0.84 (15.68\%) &  0.72 (43.71\%) &  0.71 (44.78\%) &  0.73 (37.91\%) \\
GPT3.5             &  0.85 (17.37\%) &  0.74 (45.01\%) &  0.73 (46.17\%) &  0.75 (39.29\%) \\
Mixtral 8x7B       &  0.87 (18.34\%) &  0.77 (47.50\%) &  0.77 (48.50\%) &  0.79 (42.43\%) \\
Mixtral 8x22B      &  0.92 (23.12\%) &  0.85 (52.23\%) &  0.85 (53.34\%) &  0.86 (47.42\%) \\
GPT4o              &  0.95 (25.54\%) &  0.92 (55.83\%) &  0.91 (56.80\%) &  0.92 (50.77\%) \\
GPT4-Turbo         &  0.96 (26.07\%) &  0.93 (56.30\%) &  0.92 (57.34\%) &  0.93 (51.37\%) \\
GPT4-32K           &  0.95 (25.94\%) &  0.93 (56.53\%) &  0.93 (57.56\%) &  0.93 (51.26\%) \\
\hdashline
\textbf{Inter-Model Difference} &         +3.26\% &         +7.07\% &         +7.20\% &         +6.42\% \\
\hline
\multicolumn{5}{c}{\textbf{Mistral 7B as Baseline}} \\
\hline
\rowcolor{gray!20} Mistral 7B         &    0.79 (0.00\%) &   0.50 (0.00\%) &   0.48 (0.00\%) &   0.48 (0.00\%) \\
Qwen2 1.5B         &  \cellcolor{gray!40} 0.71 (-11.47\%) &   0.51 (2.84\%) &   0.50 (4.23\%) &   0.50 (3.32\%) \\
Llama 3.1 8B       &    0.84 (6.01\%) &  0.62 (19.56\%) &  0.61 (20.82\%) &  0.61 (20.24\%) \\
GPT3.5             &    0.85 (7.88\%) &  0.66 (24.97\%) &  0.65 (26.24\%) &  0.65 (25.73\%) \\
Mixtral 8x7B       &    0.87 (8.97\%) &  0.69 (28.27\%) &  0.68 (29.30\%) &  0.68 (28.96\%) \\
Mixtral 8x22B      &   0.92 (14.30\%) &  0.82 (38.98\%) &  0.80 (40.01\%) &  0.80 (39.76\%) \\
GPT4-32K           &   0.95 (17.44\%) &  0.89 (44.26\%) &  0.88 (45.38\%) &  0.88 (45.12\%) \\
GPT4o              &   0.95 (17.00\%) &  0.90 (44.78\%) &  0.89 (45.95\%) &  0.89 (45.75\%) \\
GPT4-Turbo         &   0.96 (17.59\%) &  0.91 (45.14\%) &  0.90 (46.36\%) &  0.90 (46.08\%) \\
\hdashline
\textbf{Inter-Model Difference} &          +2.51\% &         +5.64\% &         +5.79\% &         +5.76\% \\
\hline
\multicolumn{5}{c}{\textbf{GPT3.5 as Baseline}} \\
\hline
Qwen2 1.5B         &  0.71 (-21.01\%) &  0.46 (-25.65\%) &  0.48 (-22.65\%) &  0.48 (-22.61\%) \\
Mistral 7B         &   0.79 (-8.56\%) &  0.52 (-11.02\%) &   0.55 (-6.41\%) &   0.55 (-6.51\%) \\
\rowcolor{gray!20} GPT3.5             &    0.85 (0.00\%) &    0.58 (0.00\%) &    0.59 (0.00\%) &    0.59 (0.00\%) \\
Llama 3.1 8B       &   \cellcolor{gray!40} 0.84 (-2.04\%) &    0.59 (1.79\%) &    0.60 (1.72\%) &    0.60 (1.74\%) \\
Mixtral 8x7B       &    0.87 (1.18\%) &    0.64 (8.93\%) &    0.65 (9.94\%) &    0.65 (9.57\%) \\
Mixtral 8x22B      &    0.92 (6.96\%) &   0.76 (23.69\%) &   0.79 (25.56\%) &   0.78 (25.34\%) \\
GPT4o              &    0.95 (9.89\%) &   0.82 (29.13\%) &   0.84 (29.85\%) &   0.84 (29.87\%) \\
GPT4-Turbo         &   0.96 (10.54\%) &   0.86 (32.33\%) &   0.87 (32.76\%) &   0.87 (32.68\%) \\
GPT4-32K           &   0.95 (10.38\%) &   0.86 (32.64\%) &   0.88 (33.26\%) &   0.88 (33.19\%) \\
\hdashline
\textbf{Inter-Model Difference} &          +2.11\% &          +5.44\% &          +5.54\% &          +5.53\% \\
\bottomrule
\end{tabular}
}
\caption{\footnotesize{Effect of baseline in EMDM on ARC-Challenge -- The values in parentheses indicate the percentage increase in score relative to the baseline score (Mistral 7B, represented as 0\%). The negative values (marked by grey cells) show a drop in the score relative to the baseline, marked by a grey row.}}
\label{tab:arc_c_baseline_study}
\end{table}

\subsubsection{Effect of Upper and Lower Bound in Weight Optimization}
\label{apdx:landu}
When computing the weights in Eq. \ref{eqn:pairwise_weight_optimization}, we need to choose values of $L$ and $U$. In the results section, all results were computed with $L=0.1$ and $U=2.0$. Here we repeat the experiments with the same $L$ but a higher value for the upper bound, $U=5$. The results are presented in table \ref{tab:emdm-benchmarks-u5}.

\begin{table}[h!]
\centering
\resizebox{\columnwidth}{!}{
\begin{tabular}{l|rrrr}
\toprule
{} & \textbf{EM} & \textbf{EMDM (ours)} & \textbf{Answer-only} & \textbf{Unguided-only} \\
\hline
\multicolumn{5}{c}{\textbf{ARC-Challenge}} \\
\hline
Mistral 7B         &   0.79 (0.00\%) &   0.40 (0.00\%) &   0.39 (0.00\%) &   0.45 (0.00\%) \\
Llama 3.1 8B       &   0.84 (6.01\%) &  0.48 (16.68\%) &  0.49 (20.56\%) &  0.57 (21.88\%) \\
GPT3.5             &   0.85 (7.88\%) &  0.60 (32.59\%) &  0.59 (34.42\%) &  0.62 (28.06\%) \\
Mixtral 8x7B       &   0.87 (8.97\%) &  0.62 (35.35\%) &  0.62 (37.72\%) &  0.66 (31.65\%) \\
Mixtral 8x22B      &  0.92 (14.30\%) &  0.78 (48.03\%) &  0.78 (50.27\%) &  0.79 (42.96\%) \\
GPT4-32K           &  0.95 (17.44\%) &  0.90 (55.00\%) &  0.90 (57.05\%) &  0.87 (48.45\%) \\
GPT4o              &  0.95 (17.00\%) &  0.90 (55.16\%) &  0.90 (56.97\%) &  0.88 (49.17\%) \\
GPT4-Turbo         &  0.96 (17.59\%) &  0.91 (55.82\%) &  0.92 (57.73\%) &  0.89 (49.41\%) \\
\textbf{Inter-Model Difference} &         +2.51\% &         +7.97\% &         +8.25\% &         +7.06\% \\
\hline
\multicolumn{5}{c}{\textbf{TruthfulQA}} \\
\hline
Mistral 7B         &   0.52 (0.00\%) &   0.27 (0.00\%) &   0.27 (0.00\%) &   0.35 (0.00\%) \\
GPT3.5             &  0.66 (21.75\%) &  0.44 (38.80\%) &  0.44 (39.14\%) &  0.54 (34.59\%) \\
Mixtral 8x7B       &  0.70 (26.14\%) &  0.46 (40.21\%) &  0.45 (40.20\%) &  0.57 (38.16\%) \\
Llama 3.1 8B       &  0.69 (25.09\%) &  0.50 (45.68\%) &  0.49 (45.31\%) &  0.59 (40.25\%) \\
Mixtral 8x22B      &  0.78 (33.70\%) &  0.64 (57.42\%) &  0.63 (57.44\%) &  0.71 (50.41\%) \\
GPT4o              &  0.82 (36.50\%) &  0.66 (58.90\%) &  0.66 (59.15\%) &  0.75 (52.86\%) \\
GPT4-Turbo         &  0.85 (39.16\%) &  0.76 (64.11\%) &  0.76 (64.44\%) &  0.79 (55.68\%) \\
GPT4-32K           &  0.88 (41.20\%) &  0.77 (64.81\%) &  0.77 (64.98\%) &  0.82 (57.38\%) \\
\textbf{Inter-Model Difference} &         +5.89\% &         +9.26\% &         +9.28\% &         +8.20\% \\
\hline
\multicolumn{5}{c}{\textbf{GSM8K}} \\
\hline
Mistral 7B         &   0.49 (0.00\%) &   0.28 (0.00\%) &   0.54 (0.00\%) &   0.32 (0.00\%) \\
GPT3.5             &  0.58 (15.03\%) &  0.45 (37.20\%) &  0.53 (-1.94\%) &  0.48 (34.50\%) \\
Mixtral 8x7B       &  0.69 (28.34\%) &  0.53 (46.41\%) &  0.60 (11.02\%) &  0.56 (43.90\%) \\
Llama 3.1 8B       &  0.77 (36.02\%) &  0.62 (54.24\%) &  0.81 (33.77\%) &  0.67 (52.53\%) \\
Mixtral 8x22B      &  0.87 (43.43\%) &  0.78 (64.12\%) &  0.87 (38.20\%) &  0.81 (60.92\%) \\
GPT4-32K           &  0.93 (47.20\%) &  0.88 (67.84\%) &  0.96 (43.94\%) &  0.90 (64.61\%) \\
GPT4o              &  0.94 (47.58\%) &  0.89 (68.29\%) &  0.97 (45.00\%) &  0.91 (65.05\%) \\
GPT4-Turbo         &  0.94 (47.58\%) &  0.89 (68.48\%) &  0.93 (42.22\%) &  0.91 (65.04\%) \\
\textbf{Inter-Model Difference} &         +6.85\% &         +9.78\% &         +7.50\% &         +9.29\% \\
\hline
\multicolumn{5}{c}{\textbf{MMLU: All Subjects}} \\
\hline
Mistral 7B         &   0.67 (0.00\%) &   0.66 (0.00\%) &   0.68 (0.00\%) &   0.69 (0.00\%) \\
Llama 3.1 8B       &   0.74 (8.30\%) &   0.71 (7.05\%) &   0.74 (8.09\%) &   0.74 (7.05\%) \\
GPT3.5             &  0.75 (10.36\%) &  0.74 (10.09\%) &  0.76 (10.82\%) &   0.74 (7.59\%) \\
Mixtral 8x7B       &  0.77 (12.06\%) &  0.77 (13.69\%) &  0.78 (12.66\%) &  0.77 (11.19\%) \\
Mixtral 8x22B      &  0.82 (17.62\%) &  0.82 (19.05\%) &  0.83 (18.01\%) &  0.83 (17.66\%) \\
GPT4-Turbo         &  0.86 (21.32\%) &  0.87 (23.82\%) &  0.87 (21.56\%) &  0.87 (21.42\%) \\
GPT4-32K           &  0.85 (21.04\%) &  0.87 (23.90\%) &  0.87 (21.53\%) &  0.87 (21.03\%) \\
GPT4o              &  0.88 (23.28\%) &  0.89 (25.64\%) &  0.89 (23.73\%) &  0.89 (23.44\%) \\
\textbf{Inter-Model Difference} &         +3.33\% &         +3.66\% &         +3.39\% &         +3.35\% \\
\bottomrule
\end{tabular}
}
\caption{\footnotesize{Benchmarking Results with $U=5$ -- The values in parentheses indicate the percentage increase in score relative to the baseline score (Mistral 7B, represented as 0\%). `Inter-Model Difference' shows the average model separation (difference from the closest model).}}
\label{tab:emdm-benchmarks-u5}
\end{table}

\begin{figure}
    \centering
    \includegraphics[width=1\linewidth]{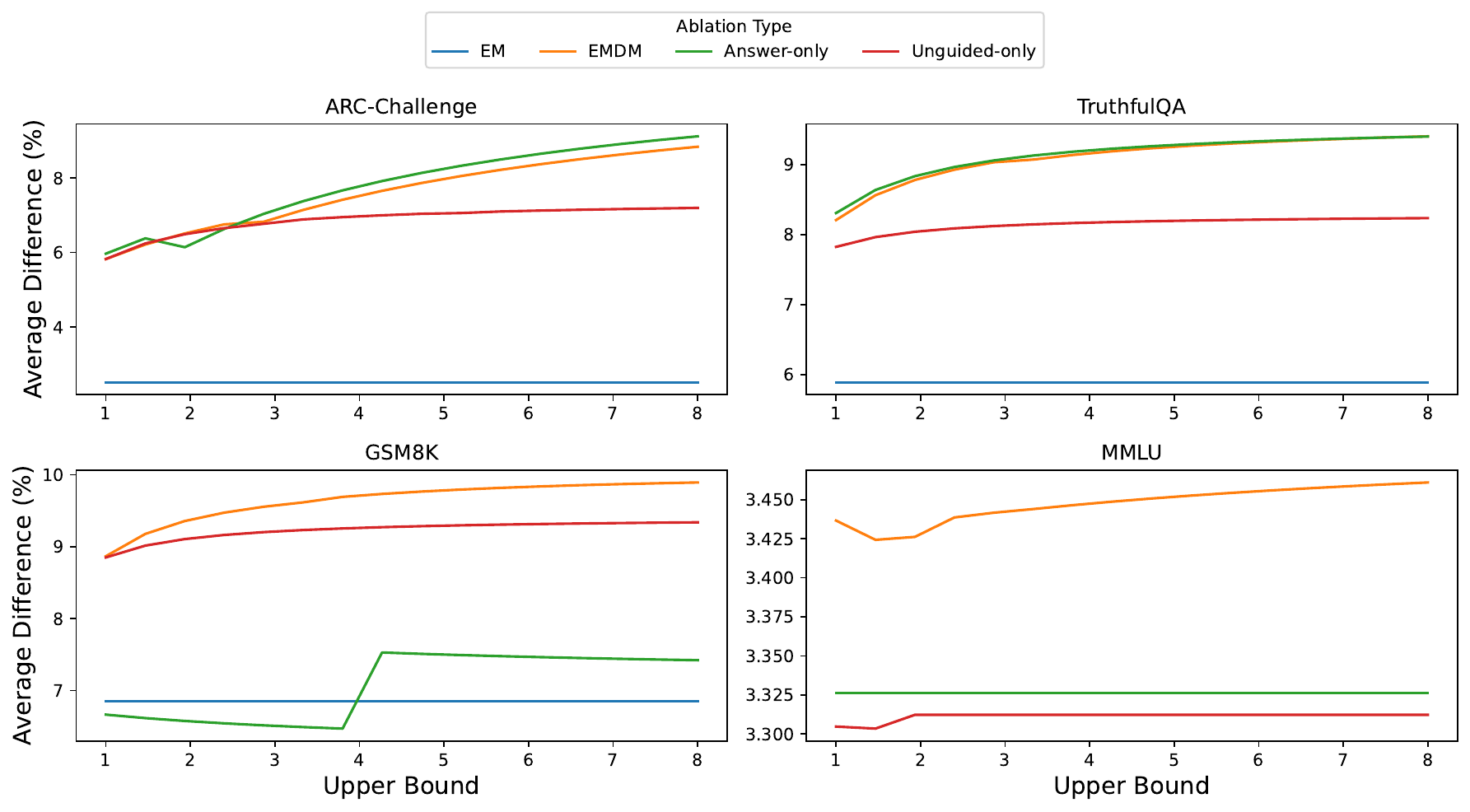}
    \caption{\footnotesize{Effect of $U$ the upper bound in weight optimization, Eqn. \ref{eqn:pairwise_weight_optimization}, on model separation.}}
    \label{fig:all_average_differences}
\end{figure}

Figure \ref{fig:all_average_differences} illustrates the overall trend of the \texttt{Inter-Model Difference} changes by increasing the upper bound limit of the weights in the weight optimization Eqn \ref{eqn:pairwise_weight_optimization}. As can be seen from table \ref{tab:emdm-benchmarks-u5} and figure \ref{fig:all_average_differences}, increasing the upper bound $U$ widens the gap and enhances model separation. However, if this value is increased excessively, it diminishes the impact of samples assigned lower weights. Given that the CoT judgment may be inaccurate and models can vary in their reasoning and ability to answer different types of questions, setting a high upper bound can be counterproductive. Nevertheless, when comparing models within the same class, such as checkpoints of the same model during training--with an earlier checkpoint as the baseline--increasing the upper bound can be beneficial.

Another key observation is the instability of the `Answer-only' method, where categorization is determined by the correctness or incorrectness of the final answer for guided and unguided responses (without using CoT judgment), resulting in 4 categories. Although, in some benchmarks, specially for larger $U$ values, `Answer-only' outperforms EMDM (using 16 categories), EMDM is always constantly improving the model separation in all the tested benchmarks.

\subsection{Full MMLU Results}
\label{apx:mmlu-results}
\begin{figure*}[t!]
\centering
\includegraphics[width=0.9\textwidth]{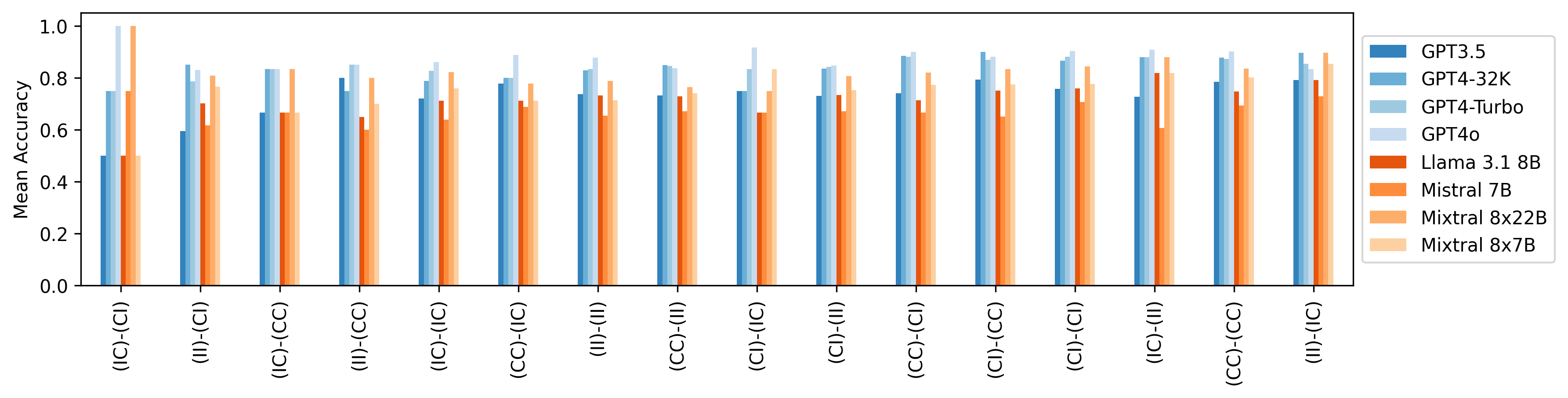}
\caption{\footnotesize{The average exact match (EM) accuracy in different sample groups of MMLU, with Mistral7B as the baseline. The groups with 0 or 1 sample are not shown.}}
\label{fig:mmlu_groups_acc}
\end{figure*}

This appendix investigates MMLU samples' behaviour in different groups assigned to them by EMDM. Figure \ref{fig:mmlu_groups_acc} shows the exact match (EM) accuracy of each category. As can be seen, different models have very similar behaviour in terms of accuracy patterns across all groups which decreases the effect of weighting in increasing the score separation. Full MMLU results across all subject classes are presented in Table \ref{tab:emdm-benchmarks-mmlu}, which presents the MMLU results based on the topic and subjects. The main observation here is the difference between humanities and other subjects.

\begin{table}[h!]
\centering
\resizebox{\columnwidth}{!}{
\begin{tabular}{l|rrrr}
\toprule
{} & \textbf{EM} & \textbf{EMDM (ours)} & \textbf{Answer-only} & \textbf{Unguided-only} \\
\hline
\multicolumn{5}{c}{\textbf{STEM}} \\
\hline
Mistral 7B         &   0.51 (0.00\%) &   0.51 (0.00\%) &   0.47 (0.00\%) &   0.53 (0.00\%) \\
GPT3.5             &   0.56 (9.78\%) &   0.55 (7.72\%) &  0.53 (11.36\%) &   0.58 (8.10\%) \\
Llama 3.1 8B       &  0.60 (14.89\%) &  0.58 (12.10\%) &  0.57 (17.82\%) &  0.61 (12.22\%) \\
Mixtral 8x7B       &  0.61 (16.38\%) &  0.59 (13.37\%) &  0.57 (17.75\%) &  0.62 (14.44\%) \\
Mixtral 8x22B      &  0.69 (26.59\%) &  0.69 (25.71\%) &  0.67 (29.30\%) &  0.70 (24.18\%) \\
GPT4-Turbo         &  0.73 (30.65\%) &  0.74 (30.99\%) &  0.70 (32.88\%) &  0.76 (30.09\%) \\
GPT4-32K           &  0.76 (33.25\%) &  0.77 (33.47\%) &  0.75 (36.94\%) &  0.78 (31.82\%) \\
GPT4o              &  0.77 (34.08\%) &  0.78 (34.41\%) &  0.76 (37.78\%) &  0.78 (31.91\%) \\
\textbf{Inter-Model Difference} &         +4.87\% &         +4.92\% &         +5.40\% &         +4.56\% \\
\hline
\multicolumn{5}{c}{\textbf{Humanities}} \\
\hline
Mistral 7B         &   0.60 (0.00\%) &   0.48 (0.00\%) &   0.51 (0.00\%) &   0.50 (0.00\%) \\
GPT3.5             &   0.63 (5.26\%) &   0.49 (0.81\%) &  0.50 (-0.24\%) &  0.49 (-1.88\%) \\
Mixtral 8x7B       &  0.67 (10.91\%) &   0.52 (7.71\%) &   0.55 (7.78\%) &   0.53 (6.40\%) \\
Llama 3.1 8B       &   0.67 (9.82\%) &  0.57 (14.80\%) &  0.58 (13.56\%) &  0.58 (13.44\%) \\
Mixtral 8x22B      &  0.76 (21.16\%) &  0.70 (31.47\%) &  0.71 (28.69\%) &  0.71 (29.77\%) \\
GPT4-Turbo         &  0.80 (24.88\%) &  0.73 (34.01\%) &  0.74 (31.95\%) &  0.75 (32.98\%) \\
GPT4o              &  0.86 (30.00\%) &  0.85 (43.58\%) &  0.84 (39.92\%) &  0.86 (41.49\%) \\
GPT4-32K           &  0.86 (29.93\%) &  0.89 (45.67\%) &  0.86 (41.50\%) &  0.88 (43.05\%) \\
\textbf{Inter-Model Difference} &         +4.29\% &         +6.52\% &         +6.92\% &         +7.18\% \\
\hline
\multicolumn{5}{c}{\textbf{Social Sciences}} \\
\hline
Mistral 7B         &   0.71 (0.00\%) &   0.69 (0.00\%) &   0.67 (0.00\%) &   0.72 (0.00\%) \\
Mixtral 8x7B       &   0.72 (1.09\%) &   0.69 (0.16\%) &   0.69 (2.90\%) &   0.72 (0.39\%) \\
GPT3.5             &   0.78 (8.89\%) &   0.75 (7.90\%) &  0.75 (10.06\%) &   0.76 (6.04\%) \\
Llama 3.1 8B       &   0.78 (9.42\%) &   0.76 (9.39\%) &  0.75 (10.70\%) &   0.77 (7.63\%) \\
GPT4-Turbo         &  0.85 (16.74\%) &  0.84 (17.89\%) &  0.82 (18.28\%) &  0.85 (16.05\%) \\
Mixtral 8x22B      &  0.87 (17.93\%) &  0.86 (19.44\%) &  0.85 (20.34\%) &  0.87 (17.46\%) \\
GPT4-32K           &  0.91 (21.64\%) &  0.91 (23.63\%) &  0.88 (23.46\%) &  0.92 (22.15\%) \\
GPT4o              &  0.92 (22.64\%) &  0.92 (24.97\%) &  0.91 (25.59\%) &  0.92 (22.57\%) \\
\textbf{Inter-Model Difference} &         +3.23\% &         +3.57\% &         +3.66\% &         +3.22\% \\
\hline
\multicolumn{5}{c}{\textbf{Others}} \\
\hline
Mistral 7B         &   0.68 (0.00\%) &   0.67 (0.00\%) &   0.65 (0.00\%) &   0.64 (0.00\%) \\
Llama 3.1 8B       &   0.74 (8.26\%) &   0.71 (5.26\%) &   0.70 (7.16\%) &   0.69 (7.39\%) \\
GPT3.5             &  0.75 (10.36\%) &   0.73 (8.22\%) &   0.72 (9.93\%) &  0.71 (10.38\%) \\
Mixtral 8x7B       &  0.77 (12.02\%) &  0.76 (11.97\%) &  0.74 (12.03\%) &  0.73 (12.09\%) \\
Mixtral 8x22B      &  0.82 (17.59\%) &  0.80 (16.57\%) &  0.79 (18.41\%) &  0.79 (19.09\%) \\
GPT4-Turbo         &  0.86 (21.29\%) &  0.86 (22.29\%) &  0.83 (22.07\%) &  0.83 (22.41\%) \\
GPT4-32K           &  0.85 (21.01\%) &  0.87 (23.26\%) &  0.83 (22.44\%) &  0.83 (22.61\%) \\
GPT4o              &  0.88 (23.25\%) &  0.88 (23.53\%) &  0.85 (23.97\%) &  0.85 (24.68\%) \\
\textbf{Inter-Model Difference} &         +3.32\% &         +3.36\% &         +3.42\% &         +3.53\% \\
\hline
\multicolumn{5}{c}{\textbf{All Subjects}} \\
\hline
Mistral 7B         &   0.67 (0.00\%) &   0.66 (0.00\%) &   0.68 (0.00\%) &   0.68 (0.00\%) \\
Llama 3.1 8B       &   0.74 (8.30\%) &   0.71 (7.12\%) &   0.74 (8.09\%) &   0.74 (7.04\%) \\
GPT3.5             &  0.75 (10.36\%) &   0.73 (9.89\%) &  0.76 (10.82\%) &   0.74 (7.84\%) \\
Mixtral 8x7B       &  0.77 (12.06\%) &  0.76 (13.45\%) &  0.78 (12.66\%) &  0.77 (11.21\%) \\
Mixtral 8x22B      &  0.82 (17.62\%) &  0.82 (19.08\%) &  0.83 (18.01\%) &  0.83 (17.61\%) \\
GPT4-32K           &  0.85 (21.04\%) &  0.86 (23.20\%) &  0.87 (21.53\%) &  0.87 (20.98\%) \\
GPT4-Turbo         &  0.86 (21.32\%) &  0.86 (23.34\%) &  0.87 (21.56\%) &  0.87 (21.38\%) \\
GPT4o              &  0.88 (23.28\%) &  0.88 (25.20\%) &  0.89 (23.73\%) &  0.89 (23.39\%) \\
\textbf{Inter-Model Difference} &         +3.33\% &         +3.60\% &         +3.39\% &         +3.34\% \\
\bottomrule
\end{tabular}
}
\caption{\footnotesize{MMLU Results -- The values in parentheses indicate the percentage increase in score relative to the baseline score (Mistral 7B, represented as 0\%).}}
\label{tab:emdm-benchmarks-mmlu}\vspace{-3mm}
\end{table}

\subsection{Example of CoT Judgement}
In this appendix, we present two randomly chosen examples from the Social Sciences subject of MMLU benchmark: 1) an example that is answered incorrectly with the unguided prompt and correctly with the guided prompt; 2) an example with the correct answer (D) but wrong reasoning in unguided set-up but correct answer and correct reasoning in guided set-up.
{\tiny
\begin{tcolorbox}[colback=white, colframe=black, title=Question]
\textbf{In order to deal with the legal issue of age discrimination, it is important for the psychologist to know that, with regard to the firing of an employee, evidence of age discrimination exists:}\\
A. only when company policy reflects adverse impact on older workers\\
B. whenever the employee's age, instead of work performance, is the basis for firing\\
C. whenever the affected employee can demonstrate substantial economic loss\\
D. only when the majority of employees in a given age group are adversely affected economically
\end{tcolorbox}

\begin{tcolorbox}[colback=red!5, colframe=red!75!black, title=Unguided Response (Incorrect)]
\textbf{Baseline Model's Chain of Thought:} \\
Option A suggests that age discrimination only exists when there is adverse impact on older workers, which is a specific condition. Options B, C, and D do not directly relate to the firing of an employee due to age discrimination, but rather focus on the employee's personal situation or economic loss, or the economic situation of a group of employees.\\
\textbf{Answer: A} \\
\textbf{CoT Judgment:} Incorrect \\
\textbf{Judge Reasoning:}
The evaluated model's reasoning is flawed because it incorrectly interprets the options provided in the question. Specifically, Option B directly addresses the core issue of age discrimination by stating that evidence of age discrimination exists 'whenever the employee's age, instead of work performance, is the basis for firing.' This directly relates to the firing of an employee due to age discrimination, contrary to the model's claim that Options B, C, and D do not directly relate to the issue. Age discrimination is fundamentally about making employment decisions based on age rather than merit or performance. Therefore, Option B is the most accurate representation of when evidence of age discrimination exists in the context of firing an employee. Options C and D focus on the consequences (economic loss) and the scale of impact (majority of employees in a given age group), which, while relevant to the broader discussion of discrimination, do not directly address the criterion for identifying age discrimination as clearly as Option B does. The model's dismissal of Option B as not directly relating to age discrimination in firing decisions is incorrect, making its chain of thought flawed.
\end{tcolorbox}

\begin{tcolorbox}[colback=green!5, colframe=green!75!black, title=Guided Response (Correct)]
\textbf{Baseline Model's Chain of Thought:} \\
Option A suggests that evidence of age discrimination exists only when company policy adversely affects older workers, which is too specific and doesn't cover all cases of age discrimination. Option C suggests that evidence exists when an affected employee can demonstrate substantial economic loss, but this doesn't necessarily mean age discrimination occurred. Option D suggests evidence exists only when the majority of employees in a given age group are adversely affected economically, which is also too specific and doesn't cover individual cases. Option B, on the other hand, states that evidence exists whenever the employee's age, instead of work performance, is the basis for firing, which aligns with the definition of age discrimination.\\
\textbf{Answer: B} \\
\textbf{CoT Judgment:} Correct \\
\textbf{Judge Reasoning:}
The reasoning provided correctly identifies why options A, C, and D do not fully capture the essence of age discrimination. Option A is too narrow because age discrimination can occur even if it's not codified in company policy. Option C is incorrect because economic loss alone does not prove age discrimination; the cause of the loss must be linked to age discrimination. Option D is also too narrow because age discrimination can affect individuals, not just majorities within age groups. The reasoning correctly identifies that Option B is the most accurate because it directly addresses the core issue of age discrimination: being fired for one's age rather than performance. This directly aligns with legal and common understandings of what constitutes age discrimination.
\end{tcolorbox}
}
{\tiny
\begin{tcolorbox}[colback=white, colframe=black, title=Question]
\textbf{The Civil Rights Act of 1964 was passed to reinforce the:}\\
A. presidential veto\\
B. system of checks and balances\\
C. states' power to challenge federal regulation\\
D. Fourteenth Amendment
\end{tcolorbox}

\begin{tcolorbox}[colback=red!5, colframe=red!75!black, title=Unguided Response (Incorrect CoT \& Correct Answer)]
\textbf{Baseline Model's Chain of Thought:} \\
The Civil Rights Act of 1964 was not passed to reinforce the presidential veto (A) as it is a law passed by Congress, not an action taken by the president. It was not passed to reinforce the system of checks and balances (B) as it is an act of legislation, not a mechanism for balancing powers. It was not passed to reinforce states' power to challenge federal regulation (C) as it is a federal law that limits states' power in certain areas. The Fourteenth Amendment (D) was a constitutional amendment, not a law passed by Congress.\\
\textbf{Answer: D} \\
\textbf{CoT Judgment:} Incorrect \\
\textbf{Judge Reasoning:}
The reasoning provided in the 'Chain of Thought' correctly identifies that options A, B, and C are not the purposes of the Civil Rights Act of 1964. However, the reasoning makes a critical error in discussing option D. The statement that the Fourteenth Amendment 'was a constitutional amendment, not a law passed by Congress' is misleading in the context of evaluating why the Civil Rights Act of 1964 was passed. The Civil Rights Act of 1964 was indeed passed to enforce the principles of the Fourteenth Amendment, specifically its provisions regarding equal protection and non-discrimination. The Fourteenth Amendment, while a constitutional amendment, provides the legal and moral foundation for subsequent legislation like the Civil Rights Act of 1964 to ensure its principles are applied. Therefore, the reasoning incorrectly dismisses the correct answer (D) by misunderstanding the relationship between constitutional amendments and legislation passed to enforce those amendments.
\end{tcolorbox}

\begin{tcolorbox}[colback=green!5, colframe=green!75!black, title=Guided Response (Correct)]
\textbf{Baseline Model's Chain of Thought:} \\
The presidential veto is a power that limits the legislative branch, not reinforces it. The system of checks and balances is a mechanism that balances the powers among the three branches, not reinforces any specific one. States' power to challenge federal regulation is limited, not reinforced, by the Civil Rights Act of 1964. The Fourteenth Amendment, on the other hand, is a constitutional amendment that reinforces the protection of civil rights.\\
\textbf{Answer: D} \\
\textbf{CoT Judgment:} Correct \\
\textbf{Judge Reasoning:}
The reasoning provided in the 'Chain of Thought' accurately identifies the purpose and effect of the Civil Rights Act of 1964 in relation to the options given. The presidential veto (Option A) is indeed a power to limit legislation, not related to the reinforcement of civil rights. The system of checks and balances (Option B) is a broad mechanism for ensuring no single branch of government has too much power, which is not directly related to the reinforcement of civil rights either. States' power to challenge federal regulation (Option C) is more about federalism dynamics than about reinforcing civil rights directly. The Fourteenth Amendment (Option D), however, is directly related to the protection of civil rights, specifically prohibiting states from denying any person 'life, liberty, or property, without due process of law' or to 'deny to any person within its jurisdiction the equal protection of the laws.' The Civil Rights Act of 1964 was indeed passed to enforce and reinforce the civil rights protections, including those outlined in the Fourteenth Amendment, by providing legal recourse against discrimination. Therefore, the reasoning steps logically support the conclusion that the Civil Rights Act of 1964 was passed to reinforce the Fourteenth Amendment.
\end{tcolorbox}
}

\subsection{Experiment Prompts}
\label{apdx:prompts}
Here is the list of prompts used in our experiments.
{\tiny
\begin{tcolorbox}[colback=gray!5, colframe=gray!75!black, title=MCQA Answering Prompt -- Elimination]
Answer the following multiple choice question. Review each option and why it can or cannot be the answer to the question, then choose the best option.
\end{tcolorbox}
\begin{tcolorbox}[colback=gray!5, colframe=gray!75!black, title=CoT Judement -- MCQA]
Evaluate the logical soundness of the reasoning provided in the 'Chain of Thought' below. Your task is to determine if the reasoning steps logically support the conclusion. Mark the 'Chain of Thought' as correct ONLY if all reasoning steps correctly lead to the conclusion. If the conclusion is unsupported by the steps or any reasoning step is flawed, mark it as incorrect. Provide a detailed justification for your judgment, specifically citing relevant parts of the reasoning. Format your evaluation in JSON, ending your response with a 'Judgment' key having the value 'correct' or 'incorrect', and ensure there is no additional text following the JSON object. The JSON object should have two keys: 'Reasoning' for your detailed justification, and 'Judgment' for your final decision.
\end{tcolorbox}
\begin{tcolorbox}[colback=gray!5, colframe=gray!75!black, title=CoT Judement -- GSM8K]
Evaluate the logical soundness of the reasoning provided in the 'Chain of Thought' below. Your task is to determine if the reasoning steps logically support the conclusion. Mark the 'Chain of Thought' as correct ONLY if all reasoning steps correctly lead to the conclusion. If the conclusion is unsupported by the steps or any reasoning step is flawed, mark it as incorrect. Provide a detailed justification for your judgment, specifically citing relevant parts of the reasoning. Format your evaluation in JSON, ending your response with a 'Judgment' key having the value 'correct' or 'incorrect', and ensure there is no additional text following the JSON object. The JSON object should have two keys: 'Reasoning' for your detailed justification, and 'Judgment' for your final decision.\\
You will be given a reference correct reasoning that must be used as an evaluation reference for the given chain of thought.
\end{tcolorbox}
}

\begin{figure}[t!]
    \centering
    \includegraphics[width=\linewidth]{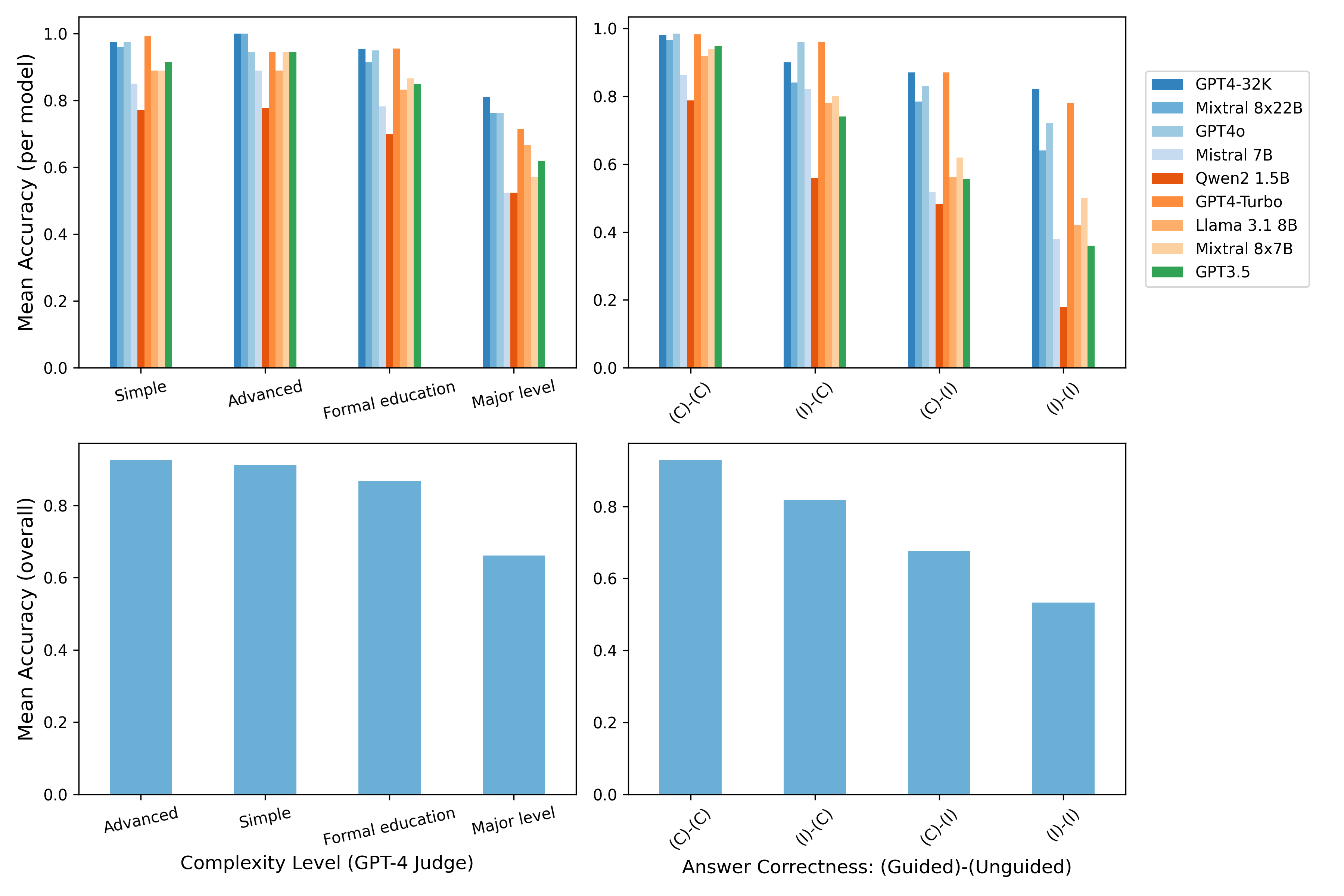}
    \caption{\footnotesize{Mean accuracy of ARC-Challenge samples from different complexity categories of FLASK and EMDM Answer-only. Top: accuracy for different models, Bottom: averaged over all models.}}
    \label{fig:complexity_emdm}
\end{figure}

\subsection{Transition Matrix for Answer-Only}
\label{apx:answer-only}
Table \ref{tab:transition-answer-only} illustrates the transition matrix of the ARC-Challenge samples in Answer-Only experiments based on the correctness of Mistral 7B answers, used in the ablation studies.

\renewcommand{\arraystretch}{2}
\begin{table}[h!]
\small
\centering
\begin{tabular}{cccc}
& & \multicolumn{2}{c}{Guided} \vspace{-3mm} \\ 
& & Correct & Incorrect \\
\midrule
\multirow{2}{*}{\rotatebox[origin=c]{90}{\parbox{1cm}{Unguided}}} 
& Correct   & 840 & 22 \\
& Incorrect & 262 & 48 \\
\bottomrule
\end{tabular}

\caption{\footnotesize{Transition matrix of Guided and Unguided prompts for ARC-Challenge, based on the final answer correctness of Mistal7B-Instruct.}}
\label{tab:transition-answer-only}
\end{table}

\subsection{FLASK Complexity vs. EMDM}
\label{apx:flask_vs_emdm}
In this appendix, we compare the effectiveness of LLM-based complexity level classification in FLASK \citep{ye2023flask} with EMDM. FLASK uses GPT-4 to assign complexity levels: 1) Simple lifestyle knowledge, 2) Advanced lifestyle knowledge, 3) Formal education knowledge, 4) Major-level knowledge, and 5) Expert-level knowledge. Figure \ref{fig:complexity_emdm} displays EMDM categories alongside FLASK's complexity assignments, with average accuracy for different models shown in the figure. We use categories from the Answer-only experiment to facilitate comparison, reducing the groups to four to match FLASK's classification. The overall mean accuracy across all models is also depicted, revealing that accuracy differences between the EMDM groups are larger than those in FLASK, indicating higher efficacy. Notably, the Simple class shows lower accuracy than the Advanced class in FLASK difficulty level classification.

\begin{figure}[t!]
    \centering
    \includegraphics[width=\linewidth]{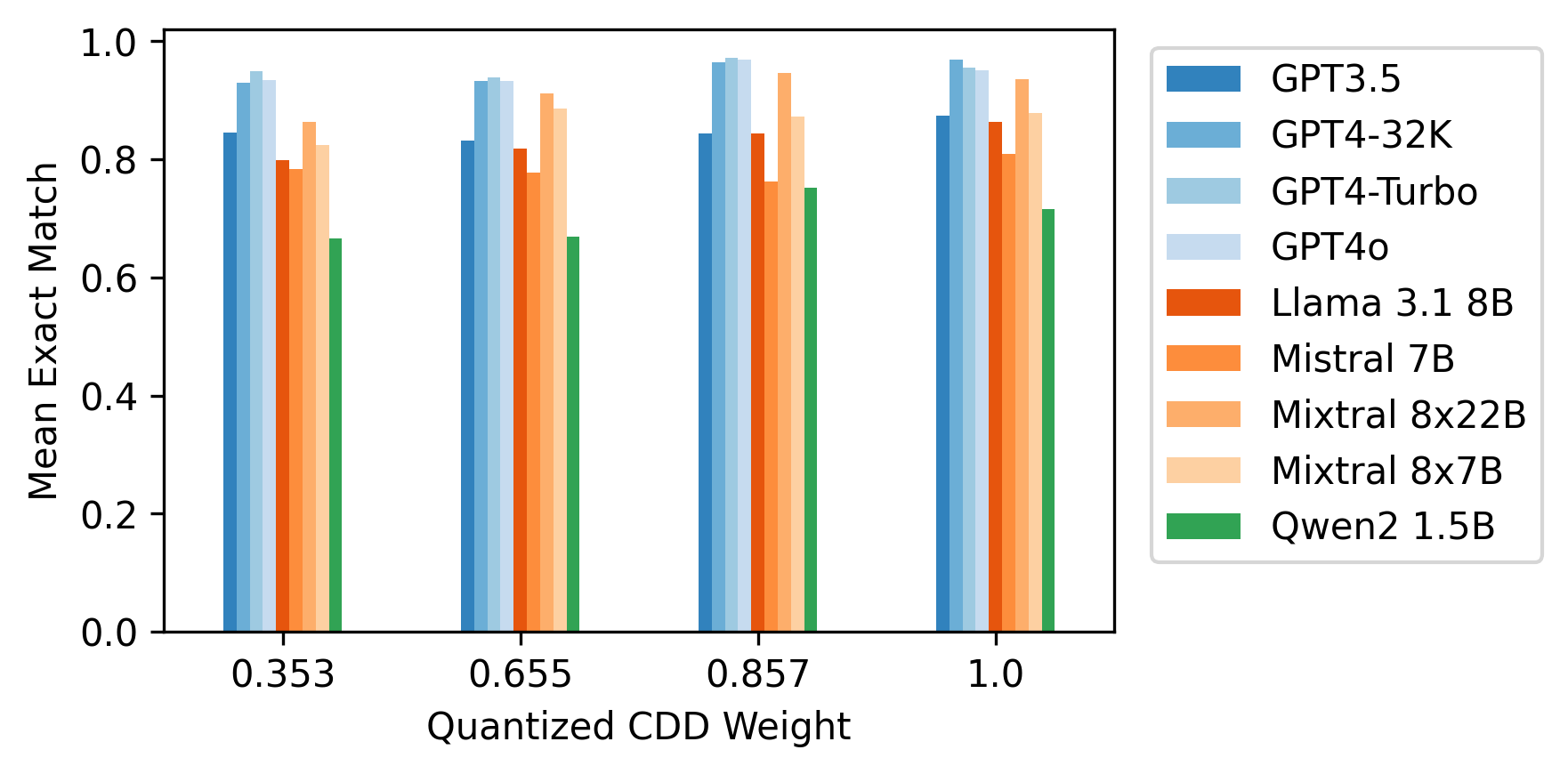}
    \caption{\footnotesize{Mean accuracy of the samples with similar CDD scores for ARC-Challenge. CDD scores and EMDM weights are calculated using Mistral 7B-Instruct as the baseline. CDD scores are quantized to four quantiles.}}
    \label{fig:cdd_score_quant}
\end{figure}

\subsection{Contamination Score vs. EMDM}
In this appendix, we want to check the accuracy of the models based on the contamination score of their samples. We group the samples based on ``Contamination Detection via output Distribution (CDD)'' \citep{dong-etal-2024-generalization} similarity of the samples. The goal of this experiment is to compare the effectiveness of EMDM grouping with contamination scores. CDD score is a metric used to assess if an LLM has memorized data from its training set. It does this by examining how ``peaked'' or repetitive the model's output is when given a specific prompt. If the outputs are very similar or identical, suggesting memorization, the CDD score will be high. A high score indicates potential data contamination, meaning the model might be relying on memorized information rather than genuinely understanding and generalizing. This is important for ensuring fair evaluations of large language model performance, especially when the training data is not publicly available. Figure \ref{fig:cdd_score_quant} shows that CDD score grouping is ineffective and all samples have the same mean accuracy. The expectation is to have higher accuracy for the samples with higher contamination scores, measured by CDD in this experiment.

EMDM samples correctly answered with flawed reasoning may be due to contamination, but the CDD experiment shows it doesn’t significantly separate model accuracy.

\subsection{Comparing Intuitive Weight Assignment with Optimized Weights}
\label{apdx:intuitive_weights}

In this section we study the effectiveness of the weight optimization in increasing the score separation. The weights are assigned based on the following logic: 1) if the answer and CoT is incorrect in any of the prompt types, the weight is 2.0 (higher bound), 2) if the answer and reasoning are both correct the weight is 0.1, and 3) anything else gets a weight on 1.0. This weighting schema are shown in table \ref{tab:emdm_intuitive_weights} for 16 categories, table \ref{tab:answer_only_ituitive_weights} for Answer-only ablation, and table \ref{tab:unguided_only_intuitive_weights} for unguided prompt only.

\begin{table}[ht]
\centering
\resizebox{0.4\columnwidth}{!}{
\begin{tabular}{@{}cc|cccc@{}}
 & & \multicolumn{4}{c}{\textbf{Guided}} \\
 & & \textbf{CC} & \textbf{CI} & \textbf{IC} & \textbf{II} \\ 
\midrule
\multirow{4}{*}{\rotatebox{90}{\textbf{Unguided}}} 
& \textbf{CC} & 0.1 & 0.1 & 0.1 & 2.0 \\
& \textbf{CI} & 0.1 & 1.0 & 1.0 & 2.0 \\
& \textbf{IC} & 0.1 & 1.0 & 1.0 & 2.0 \\
& \textbf{II} & 2.0 & 2.0 & 2.0 & 2.0 \\
\bottomrule
\end{tabular}
}
\caption{\footnotesize{Transition matrix for EMDM weights. Guided and Unguided correctness categories are shown.}}
\label{tab:emdm_intuitive_weights}
\end{table}

\begin{table}[ht]
\centering
\resizebox{0.25\columnwidth}{!}{
\begin{tabular}{@{}cc|cc@{}}
 & & \multicolumn{2}{c}{\textbf{Guided}} \\
 & & \textbf{C} & \textbf{I} \\ 
\midrule
\multirow{2}{*}{\rotatebox{90}{\textbf{Unguided}}} 
& \textbf{C} & 0.1 & 1.0 \\
& \textbf{I} & 1.0 & 2.0 \\
\bottomrule
\end{tabular}
}
\caption{\footnotesize{Transition matrix for Answer-Only weights.}}
\label{tab:answer_only_ituitive_weights}
\end{table}

\begin{table}[ht]
\centering
\resizebox{0.4\columnwidth}{!}{
\begin{tabular}{@{}c|cccc@{}}
 & \textbf{CC} & \textbf{CI} & \textbf{IC} & \textbf{II} \\ 
\midrule
\textbf{Unguided} & 0.1 & 1.0 & 1.0 & 2.0 \\
\bottomrule
\end{tabular}
}
\caption{\footnotesize{Unguided-Only weights.}}
\label{tab:unguided_only_intuitive_weights}
\end{table}

\begin{table}[h!]
\centering
\resizebox{\columnwidth}{!}{
\begin{tabular}{l|c|rrr|rrr}
\toprule
{} & {} & \multicolumn{3}{c}{\textbf{Optimized}} & \multicolumn{3}{c}{\textbf{Intuitive}} \\
\midrule
{} & \textbf{EM} & \textbf{EMDM (ours)} & \textbf{Answer-only} & \textbf{Unguided-only} & \textbf{EMDM$^*$} & \textbf{Answer-only} & \textbf{Unguided-only}\\
\hline
\multicolumn{8}{c}{\textbf{ARC-Challenge}} \\
\hline

Mistral 7B         &   0.79 (0.00\%) &   0.46 (0.00\%) &   0.52 (0.00\%) &   0.49 (0.00\%) &   0.57 (0.00\%) &   0.50 (0.00\%) &   0.57 (0.00\%) \\
Llama 3.1 8B       &   0.84 (6.01\%) &  0.55 (17.54\%) &  0.61 (13.65\%) &  0.60 (19.17\%) &  0.67 (15.67\%) &  0.62 (18.95\%) &  0.67 (15.17\%) \\
GPT3.5             &   0.85 (7.88\%) &  0.60 (24.56\%) &  0.68 (22.96\%) &  0.65 (24.94\%) &  0.71 (20.08\%) &  0.66 (24.86\%) &  0.71 (20.07\%) \\
Mixtral 8x7B       &   0.87 (8.97\%) &  0.63 (27.15\%) &  0.71 (25.58\%) &  0.68 (28.32\%) &  0.73 (22.26\%) &  0.69 (27.72\%) &  0.73 (21.93\%) \\
Mixtral 8x22B      &  0.92 (14.30\%) &  0.76 (39.66\%) &  0.83 (36.51\%) &  0.80 (39.22\%) &  0.83 (31.94\%) &  0.82 (38.74\%) &  0.83 (31.64\%) \\
GPT4-32K           &  0.95 (17.44\%) &  0.83 (44.99\%) &  0.92 (42.90\%) &  0.88 (44.72\%) &  0.90 (37.10\%) &  0.89 (44.12\%) &  0.90 (36.75\%) \\
GPT4-Turbo         &  0.96 (17.59\%) &  0.85 (46.26\%) &  0.93 (43.52\%) &  0.89 (45.62\%) &  0.91 (37.99\%) &  0.91 (44.99\%) &  0.91 (37.66\%) \\
GPT4o              &  0.95 (17.00\%) &  0.86 (\textbf{46.81}\%) &  0.92 (42.72\%) &  0.89 (45.37\%) & 0.91 (37.99\%) &  0.90 (44.64\%) &  0.90 (37.38\%) \\
\hdashline
\textbf{Inter-Model Difference} &         +2.51\% &         \textbf{+6.69}\% &         +6.22\% &         +6.52\%  &         +5.43\% &         +6.43\% &         +5.38\% \\
\hline
\multicolumn{8}{c}{\textbf{TruthfulQA}} \\
\hline
Mistral 7B         &   0.52 (0.00\%) &   0.30 (0.00\%) &   0.30 (0.00\%) &   0.36 (0.00\%) &   0.42 (0.00\%) &   0.35 (0.00\%) &   0.41 (0.00\%) \\
GPT3.5             &  0.66 (21.75\%) &  0.47 (35.94\%) &  0.47 (36.40\%) &  0.54 (33.68\%) &  0.59 (29.22\%) &  0.53 (33.37\%) &  0.58 (29.59\%) \\
Mixtral 8x7B       &  0.70 (26.14\%) &  0.48 (37.71\%) &  0.48 (37.91\%) &  0.58 (37.35\%) &  0.62 (32.82\%) &  0.56 (36.36\%) &  0.61 (33.04\%) \\
Llama 3.1 8B       &  0.69 (25.09\%) &  0.52 (42.51\%) &  0.52 (42.29\%) &  0.59 (39.25\%) &  0.63 (33.64\%) &  0.58 (38.75\%) &  0.63 (34.29\%) \\
Mixtral 8x22B      &  0.78 (33.70\%) &  0.66 (54.08\%) &  0.65 (54.27\%) &  0.71 (49.36\%) &  0.74 (43.09\%) &  0.70 (49.52\%) &  0.73 (43.62\%) \\
GPT4o              &  0.82 (36.50\%) &  0.68 (55.64\%) &  0.68 (56.12\%) &  0.75 (51.78\%) &  0.78 (46.12\%) &  0.74 (52.01\%) &  0.77 (46.51\%) \\
GPT4-Turbo         &  0.85 (39.16\%) &  0.77 (60.95\%) &  0.77 (61.33\%) &  0.80 (54.70\%) &  0.82 (48.70\%) &  0.80 (55.70\%) &  0.81 (49.17\%) \\
GPT4-32K           &  0.88 (41.20\%) &  0.79 (61.65\%) &  0.78 (\textbf{62.01}\%) &  0.83 (56.35\%) &  0.85 (50.58\%) &  0.82 (56.85\%) &  0.84 (51.21\%) \\
\hdashline
\textbf{Inter-Model Difference} &         +5.89\% &         +8.81\% &         \textbf{+8.86}\% &         +8.05\%    &         +7.23\% &         +8.12\% &         +7.32\%\\
\hline
\multicolumn{8}{c}{\textbf{GSM8K}} \\
\hline
Mistral 7B         &   0.49 (0.00\%) &   0.31 (0.00\%) &   0.52 (0.00\%) &   0.33 (0.00\%) &   0.43 (0.00\%) &   0.47 (0.00\%) &   0.41 (0.00\%) \\
GPT3.5             &  0.58 (15.03\%) &  0.46 (32.69\%) &   0.55 (5.24\%) &  0.49 (32.78\%) &  0.55 (22.51\%) &  0.58 (18.41\%) &  0.54 (23.49\%) \\
Mixtral 8x7B       &  0.69 (28.34\%) &  0.54 (42.88\%) &  0.64 (18.49\%) &  0.57 (42.59\%) &  0.64 (33.35\%) &  0.67 (29.72\%) &  0.63 (34.14\%) \\
Llama 3.1 8B       &  0.77 (36.02\%) &  0.64 (51.56\%) &  0.79 (34.64\%) &  0.67 (51.22\%) &  0.73 (41.48\%) &  0.76 (37.43\%) &  0.72 (42.65\%) \\
Mixtral 8x22B      &  0.87 (43.43\%) &  0.79 (61.09\%) &  0.87 (40.30\%) &  0.82 (59.62\%) &  0.85 (49.61\%) &  0.87 (45.29\%) &  0.84 (50.85\%) \\
GPT4-32K           &  0.93 (47.20\%) &  0.88 (65.07\%) &  0.95 (45.22\%) &  0.90 (63.37\%) &  0.92 (53.44\%) &  0.93 (49.21\%) &  0.92 (54.73\%) \\
GPT4o              &  0.94 (47.58\%) &  0.90 (65.57\%) &  0.96 (46.01\%) &  0.91 (63.81\%) &  0.93 (53.82\%) &  0.93 (49.29\%) &  0.92 (55.12\%) \\
GPT4-Turbo         &  0.94 (47.58\%) &  0.90 (\textbf{65.64}\%) &  0.93 (44.38\%) &  0.91 (63.78\%) &  0.93 (53.82\%) &  0.94 (49.41\%) &  0.92 (55.12\%) \\
\hdashline
\textbf{Inter-Model Difference} &         +6.85\% &         \textbf{+9.38}\% &         +6.57\% &         +9.12\%    &         +7.69\% &         +7.06\% &         +7.93\%\\
\hline
\multicolumn{8}{c}{\textbf{MMLU: All Subjects}} \\
\hline
Mistral 7B         &   0.67 (0.00\%) &   0.66 (0.00\%) &   0.68 (0.00\%) &   0.68 (0.00\%) &   0.67 (0.00\%) &   0.66 (0.00\%) &   0.67 (0.00\%) \\
Llama 3.1 8B       &   0.74 (8.30\%) &   0.71 (7.12\%) &   0.74 (8.09\%) &   0.74 (7.04\%) &   0.73 (8.85\%) &   0.73 (9.47\%) &   0.73 (8.88\%) \\
GPT3.5             &  0.75 (10.36\%) &   0.73 (9.89\%) &  0.76 (10.82\%) &   0.74 (7.84\%) &   0.74 (9.21\%) &  0.73 (10.30\%) &   0.74 (9.34\%) \\
Mixtral 8x7B       &  0.77 (12.06\%) &  0.76 (13.45\%) &  0.78 (12.66\%) &  0.77 (11.21\%) &  0.75 (11.08\%) &  0.75 (11.72\%) &  0.75 (11.01\%) \\
Mixtral 8x22B      &  0.82 (17.62\%) &  0.82 (19.08\%) &  0.83 (18.01\%) &  0.83 (17.61\%) &  0.81 (17.17\%) &  0.81 (18.25\%) &  0.81 (17.16\%) \\
GPT4-32K           &  0.85 (21.04\%) &  0.86 (23.20\%) &  0.87 (21.53\%) &  0.87 (20.98\%) &  0.84 (20.68\%) &  0.83 (20.51\%) &  0.84 (20.70\%) \\
GPT4-Turbo         &  0.86 (21.32\%) &  0.86 (23.34\%) &  0.87 (21.56\%) &  0.87 (21.38\%) &  0.85 (21.11\%) &  0.84 (21.59\%) &  0.85 (21.29\%) \\
GPT4o              &  0.88 (23.28\%) &  0.88 (\textbf{25.20}\%) &  0.89 (23.73\%) &  0.89 (23.39\%) &  0.87 (22.80\%) &  0.86 (23.69\%) &  0.87 (23.00\%) \\
\hdashline
\textbf{Inter-Model Difference} &         +3.33\% &         \textbf{+3.60}\% &         +3.39\% &         +3.34\%  &         +3.26\% &         +3.38\% &         +3.29\% \\
\bottomrule
\end{tabular}
}
\caption{\footnotesize{Benchmarking Results -- The values in parentheses indicate the percentage increase in score relative to the baseline, represented as 0\%. `Inter-Model Difference' shows the average model separation (difference from the closest model).}}
\label{tab:emdm-benchmarks-intuitive_vs_optimized}
\end{table}

In addition to the above weighting schema, we also tested another method: the weights of 2.0 and 0.1 are assigned similar to the above, but wherever weight is 1.0 they are scaled by the ratio of the samples in each category $\dfrac{\text{Group Count}}{\text{Total Count}}$. The results are shown in table \ref{tab:emdm-benchmarks-intuitive_vs_optimized_countnorm}.

\begin{table}[h!]
\centering
\resizebox{\columnwidth}{!}{
\begin{tabular}{l|c|rrr|rrr}
\toprule
{} & {} & \multicolumn{3}{c}{\textbf{Optimized}} & \multicolumn{3}{c}{\textbf{Intuitive}} \\
\midrule
{} & \textbf{EM} & \textbf{EMDM (ours)} & \textbf{Answer-only} & \textbf{Unguided-only} & \textbf{EMDM$^*$} & \textbf{Answer-only} & \textbf{Unguided-only}\\
\hline
\multicolumn{8}{c}{\textbf{ARC-Challenge}} \\
\hline

Mistral 7B         &   0.79 (0.00\%) &   0.46 (0.00\%) &   0.52 (0.00\%) &   0.49 (0.00\%) &   0.49 (0.00\%) &   0.52 (0.00\%) &   0.50 (0.00\%)    \\
Llama 3.1 8B       &   0.84 (6.01\%) &  0.55 (17.54\%) &  0.61 (13.65\%) &  0.60 (19.17\%) &  0.61 (19.92\%) &  0.61 (15.13\%) &  0.61 (18.47\%)   \\
GPT3.5             &   0.85 (7.88\%) &  0.60 (24.56\%) &  0.68 (22.96\%) &  0.65 (24.94\%) &  0.65 (24.65\%) &  0.68 (23.70\%) &  0.66 (24.18\%)   \\
Mixtral 8x7B       &   0.87 (8.97\%) &  0.63 (27.15\%) &  0.71 (25.58\%) &  0.68 (28.32\%) &  0.68 (28.54\%) &  0.70 (26.43\%) &  0.68 (27.41\%)   \\
Mixtral 8x22B      &  0.92 (14.30\%) &  0.76 (39.66\%) &  0.83 (36.51\%) &  0.80 (39.22\%) &  0.81 (39.56\%) &  0.82 (37.27\%) &  0.80 (38.21\%)   \\
GPT4-32K           &  0.95 (17.44\%) &  0.83 (44.99\%) &  0.92 (42.90\%) &  0.88 (44.72\%) &  0.89 (44.84\%) &  0.91 (43.50\%) &  0.88 (43.61\%)   \\
GPT4-Turbo         &  0.96 (17.59\%) &  0.85 (46.26\%) &  0.93 (43.52\%) &  0.89 (45.62\%) &  0.90 (45.71\%) &  0.92 (44.20\%) &  0.90 (44.51\%)   \\
GPT4o              &  0.95 (17.00\%) &  0.86 (\textbf{46.81}\%) &  0.92 (42.72\%) &  0.89 (45.37\%) &  0.90 (45.71\%) &  0.91 (43.50\%) &  0.89 (44.27\%)  \\
\hdashline
\textbf{Inter-Model Difference} &         +2.51\% &         \textbf{+6.69}\% &         +6.22\% &         +6.52\%  &   +6.53\% &         +6.31\% &         +6.36\% \\
\hline
\multicolumn{8}{c}{\textbf{TruthfulQA}} \\
\hline
Mistral 7B         &   0.52 (0.00\%) &   0.30 (0.00\%) &   0.30 (0.00\%) &   0.36 (0.00\%) &   0.38 (0.00\%) &   0.31 (0.00\%) &   0.37 (0.00\%)    \\
GPT3.5             &  0.66 (21.75\%) &  0.47 (35.94\%) &  0.47 (36.40\%) &  0.54 (33.68\%) &  0.56 (32.52\%) &  0.48 (35.88\%) &  0.55 (32.94\%)   \\
Mixtral 8x7B       &  0.70 (26.14\%) &  0.48 (37.71\%) &  0.48 (37.91\%) &  0.58 (37.35\%) &  0.59 (35.98\%) &  0.50 (37.88\%) &  0.58 (36.50\%)   \\
Llama 3.1 8B       &  0.69 (25.09\%) &  0.52 (42.51\%) &  0.52 (42.29\%) &  0.59 (39.25\%) &  0.60 (37.27\%) &  0.53 (41.74\%) &  0.60 (38.33\%)   \\
Mixtral 8x22B      &  0.78 (33.70\%) &  0.66 (54.08\%) &  0.65 (54.27\%) &  0.71 (49.36\%) &  0.72 (47.64\%) &  0.66 (53.34\%) &  0.71 (48.33\%)  \\
GPT4o              &  0.82 (36.50\%) &  0.68 (55.64\%) &  0.68 (56.12\%) &  0.75 (51.78\%) &  0.76 (50.51\%) &  0.69 (55.26\%) &  0.75 (50.92\%)   \\
GPT4-Turbo         &  0.85 (39.16\%) &  0.77 (60.95\%) &  0.77 (61.33\%) &  0.80 (54.70\%) &  0.81 (53.16\%) &  0.77 (60.10\%) &  0.80 (53.65\%)   \\
GPT4-32K           &  0.88 (41.20\%) &  0.79 (61.65\%) &  0.78 (\textbf{62.01}\%) &  0.83 (56.35\%)  &  0.84 (54.71\%) &  0.79 (60.93\%) &  0.83 (55.50\%)   \\
\hdashline
\textbf{Inter-Model Difference} &         +5.89\% &         +8.81\% &         \textbf{+8.86}\%  &         +8.05\%     &    +7.82\% &         +8.70\% &         +7.93\%   \\
\hline
\multicolumn{8}{c}{\textbf{GSM8K}} \\
\hline
Mistral 7B         &   0.49 (0.00\%) &   0.31 (0.00\%) &   0.52 (0.00\%) &   0.33 (0.00\%) &   0.38 (0.00\%) &   0.47 (0.00\%) &   0.36 (0.00\%)   \\
GPT3.5             &  0.58 (15.03\%) &  0.46 (32.69\%) &   0.55 (5.24\%) &  0.49 (32.78\%) &  0.53 (27.29\%) &  0.58 (18.79\%) &  0.51 (29.79\%)   \\
Mixtral 8x7B       &  0.69 (28.34\%) &  0.54 (42.88\%) &  0.64 (18.49\%) &  0.57 (42.59\%) &  0.61 (37.55\%) &  0.67 (29.80\%) &  0.59 (39.69\%)   \\
Llama 3.1 8B       &  0.77 (36.02\%) &  0.64 (51.56\%) &  0.79 (34.64\%) &  0.67 (51.22\%) &  0.70 (45.67\%) &  0.75 (37.33\%) &  0.69 (48.32\%)   \\
Mixtral 8x22B      &  0.87 (43.43\%) &  0.79 (61.09\%) &  0.87 (40.30\%) &  0.82 (59.62\%) &  0.84 (54.25\%) &  0.86 (45.36\%) &  0.82 (56.73\%)  \\
GPT4-32K           &  0.93 (47.20\%) &  0.88 (65.07\%) &  0.95 (45.22\%) &  0.90 (63.37\%) &  0.91 (58.03\%) &  0.93 (49.31\%) &  0.90 (60.51\%)  \\
GPT4o              &  0.94 (47.58\%) &  0.90 (65.57\%) &  0.96 (46.01\%) &  0.91 (63.78\%) &  0.92 (58.39\%) &  0.93 (49.34\%) &  0.91 (60.94\%)  \\
GPT4-Turbo         &  0.94 (47.58\%) &  0.90 (\textbf{65.64}\%) &  0.93 (44.38\%) &  0.91 (63.78\%) &  0.92 (58.40\%) &  0.94 (49.52\%) &  0.91 (60.95\%)  \\
\hdashline
\textbf{Inter-Model Difference} &         +6.85\% &         \textbf{+9.38}\% &         +6.57\% &         +9.12\%    &      +8.35\% &         +7.07\% &         +8.71\%   \\
\hline
\multicolumn{8}{c}{\textbf{MMLU: All Subjects}} \\
\hline
Mistral 7B         &   0.67 (0.00\%) &   0.66 (0.00\%) &   0.68 (0.00\%) &   0.68 (0.00\%) &   0.67 (0.00\%) &   0.66 (0.00\%) &   0.67 (0.00\%)    \\
Llama 3.1 8B       &   0.74 (8.30\%) &   0.71 (7.12\%) &   0.74 (8.09\%) &   0.74 (7.04\%) &   0.73 (9.37\%) &  0.73 (10.18\%) &   0.73 (9.31\%)   \\
GPT3.5             &  0.75 (10.36\%) &   0.73 (9.89\%) &  0.76 (10.82\%) &   0.74 (7.84\%) &   0.73 (9.40\%) &  0.74 (11.08\%) &   0.74 (10.59\%)    \\
Mixtral 8x7B       &  0.77 (12.06\%) &  0.76 (13.45\%) &  0.78 (12.66\%) &  0.77 (11.21\%) &  0.75 (10.93\%) &  0.75 (12.14\%) &  0.74 (10.61\%)   \\
Mixtral 8x22B      &  0.82 (17.62\%) &  0.82 (19.08\%) &  0.83 (18.01\%) &  0.83 (17.61\%) &  0.80 (16.87\%) &  0.81 (19.06\%) &  0.80 (16.68\%)   \\
GPT4-32K           &  0.85 (21.04\%) &  0.86 (23.20\%) &  0.87 (21.53\%) &  0.87 (20.98\%) &  0.84 (20.95\%) &  0.83 (20.58\%) &  0.84 (20.92\%)   \\
GPT4-Turbo         &  0.86 (21.32\%) &  0.86 (23.34\%) &  0.87 (21.56\%) &  0.87 (21.38\%) &  0.84 (21.03\%) &  0.84 (22.06\%) &  0.85 (21.32\%)   \\
GPT4o              &  0.88 (23.28\%) &  0.88 (\textbf{25.20}\%) &  0.89 (23.73\%) &  0.89 (23.39\%) &  0.86 (22.42\%) &  0.87 (24.62\%) &  0.86 (22.78\%)   \\
\hdashline
\textbf{Inter-Model Difference} &         +3.33\% &         \textbf{+3.60}\% &         +3.39\% &         +3.34\%  &      +3.20\% &         +3.52\% &         +3.25\% \\
\bottomrule
\end{tabular}
}
\caption{\footnotesize{Benchmarking Results -- The values in parentheses indicate the percentage increase in score relative to the baseline, represented as 0\%. `Inter-Model Difference' shows the average model separation (difference from the closest model).}}
\label{tab:emdm-benchmarks-intuitive_vs_optimized_countnorm}
\end{table}

As can be seen from both tables \ref{tab:emdm-benchmarks-intuitive_vs_optimized_countnorm} and \ref{tab:emdm-benchmarks-intuitive_vs_optimized}, EMDM with optimized weights achieve the highest separation. The Answer-only results are better with the intuitive weights in some benchmarks, due to lower degree of freedom in its optimization. This can be improved by changing the weight of the terms in the optimization objective in Eq. \ref{eqn:pairwise_weight_optimization}.

\subsection{Computation Costs}
We used Azure OpenAI\footnote{\url{https://azure.microsoft.com/en-us/products/ai-services/openai-service}} endpoint with API-version of ``2024-02-15-preview'' for OpenAI models. Qwen, Mistral, Llama 3.1, and Mixtral models were hosted on a Node with 8xH100 GPUs served with vLLM\footnote{\url{https://github.com/vllm-project/vllm}}.

\subsection{Weight Optimization as a Linear Programming Problem}
\label{apdx:linear_programming}

Given the objective function:

\begin{equation*}
\small
\begin{split}
&\hat{w} = \mathop{\mathrm{argmax}}_{\substack{w=\{w_{g_k}\}_{k=1}^N \\ L \leq w_{g_k} \leq U}} 
-\sum_{g_k \in G} w_{g_k}  
+ \sum_{g_k \in G} \frac{w_{g_k}}{n_{g_k} \cdot \binom{|M|}{2}} \\
&\quad \times \sum_{\substack{(A, B) \in M \\ A \neq B \\ A, B \neq R}} \sum_{i \in g_k} \left| Score^A(x_i) - Score^B(x_i) \right|
\end{split}
\end{equation*}

Let's define:

\begin{equation*}
\small
\begin{split}
& D_{g_k} = \frac{\left(\frac{|M|}{2}\right)}{n_{g_k}} \sum_{\substack{(A, B) \in M \\ A \neq B \\ A, B \neq R}} \\
& \sum_{i \in g_k} \left| \operatorname{Score}^A(x_i) - \operatorname{Score}^B(x_i) \right|. 
\end{split}
\end{equation*}

Using the above definition, we can reformulate the objective function as:

\begin{equation*}
\small
\min_{\mathbf{w}} \quad \sum_{k=1}^N -w_{g_k} D_{g_k} + w_{g_k}.
\end{equation*}

This can be written in the standard Linear Programming (LP) form:

\begin{equation*}
\begin{aligned}
&\min_{\mathbf{w}} \quad \mathbf{c}^T \mathbf{w}, \\
&\text{subject to} \quad L \leq w_{g_k} \leq U, \quad \forall k = 1, 2, \dots, N, \\
&\text{where} \quad \mathbf{c} = \{1 - D_{g_k}\}_{k=1}^N.
\end{aligned}
\end{equation*}

\end{document}